\documentclass[10pt,twocolumn,letterpaper]{article}

\usepackage[pagenumbers]{cvpr}

\usepackage{cuted} %< for strip environment

\usepackage{currfile} %< for \currfilebase macro

\usepackage{caption} %< for "\captionof" commands (teaser)
\definecolor{ectorange}{HTML}{EB6834}
\usepackage{booktabs,multirow,colortbl}
\usepackage{needspace}

\definecolor{cvprblue}{rgb}{0.21,0.49,0.74}
\usepackage[pagebackref,breaklinks,colorlinks,allcolors=cvprblue]{hyperref}
\newcommand{\suppref}[1]{\hyperref[#1]{Supplement~\ref*{#1}}}

\title{Speak When You See It: Proactive Task Assistants Should Wait for Evidence, Not the Clock}
\author{Bohe Chen\\
  {\normalsize Beijing University of Posts and Telecommunications, Beijing, China}\\
  {\normalsize Queen Mary University of London, London, UK}\\
  {\normalsize\href{mailto:bhchen@bupt.edu.cn}{\texttt{bhchen@bupt.edu.cn}}}}
\hypersetup{pdfauthor={Bohe Chen},pdftitle={Speak When You See It: Proactive Task Assistants Should Wait for Evidence, Not the Clock}}

\begin{document}
\twocolumn[{%
\renewcommand\twocolumn[1][]{#1}%
\maketitle
\begin{center}
  \centering
  \captionsetup{type=figure}
  \includegraphics[width=\linewidth]{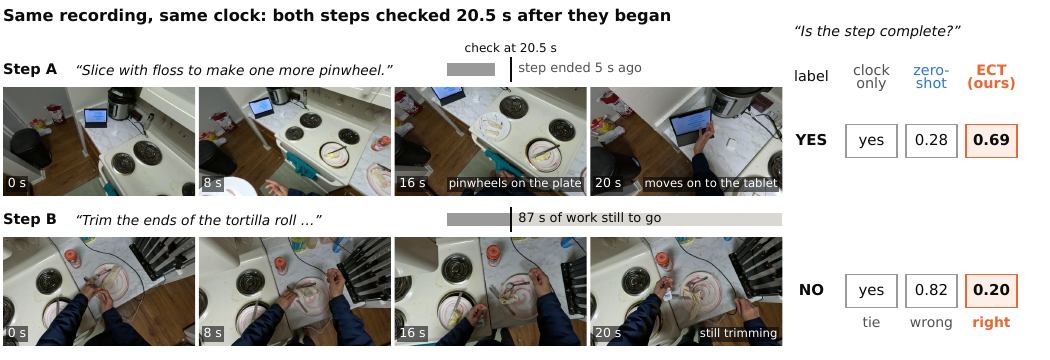}
  \captionof{figure}{\textbf{Same clock, different evidence.}
  Two steps of one QICD test recording, each checked 20.5\,s after it began: step~A has just ended, step~B has 87\,s to go.
  A policy of elapsed time alone ties the two checks, the zero-shot model ranks them in the wrong order, and ECT ranks them correctly (ECT probabilities average four seeds). The example is the median pair under a fixed selection rule (\suppref{sup:pairs}).
  Across 931 time-matched test pairs, elapsed time alone scores 0.50 by construction (0.65 and 0.54 with the instruction's typical duration), the zero-shot model scores 0.66 on completion and 0.59 on mistake questions, and ECT scores 0.90--0.92 and 0.75--0.77.}
  \label{fig:teaser}
\end{center}%
}]
\begingroup
\setlength{\textfloatsep}{12pt plus 2pt minus 2pt}
\setlength{\dbltextfloatsep}{14pt plus 2pt minus 2pt}
\begin{abstract}
Proactive assistants that guide people through physical tasks must decide, every few seconds, whether a step is complete and whether a mistake has occurred.
We find that two benchmarks for this task largely reward \emph{knowing when} rather than \emph{seeing what}: a timer that never looks at the video, raising an alarm at 25\,s and confirming each step at 30\,s, reaches a mistake F1 of 0.373 on Ego-MC-Bench, above every published result (at most 0.27).
We formalize this as the video-blind frontier, the best trade-off between completion accuracy and mistake F1 that policies reach from the elapsed time, the instruction and training statistics alone: every published per-step result on Ego-MC-Bench lies inside it, and so do all twelve zero-shot models we run on QICD.
To move beyond it, we introduce Evidence-Contrastive Training (ECT): fine-tune on the benchmark's own questions, and pair its positives with negatives from the same recording at the same elapsed time, so that elapsed time cannot separate the two sides of a pair.
Trained on 209 QICD videos in about 15 GPU-hours, a 2B model with ECT crosses the frontier on both benchmarks in all four seeds; transferred to Ego-MC-Bench without adaptation, it confirms 59\% of mistake-free steps in time, against 30\% for a published model with the same backbone trained on about 25k synthetic counterfactuals and real recordings.
Its decisions follow the video beyond what timing explains: at equal elapsed time it tells a finished step from an unfinished one in 90--92\% of test pairs, where the instruction's typical duration reaches 65\%, rescaling the time stamps between the frames leaves it essentially unchanged, and swapping in another recording, of the same recipe where one exists, cuts its completion accuracy by two thirds.
The code and model weights will be publicly released.
\end{abstract}

\section{Introduction}
\label{sec:intro}

A proactive assistant watches a person cook, assemble or repair through a head-mounted camera, and must decide every few seconds whether the current step is complete and whether something has gone wrong~\cite{bhattacharyya2025livemamba,bhattacharyya2026streaming,wang2023holoassist,liu2026guideme}.
Speaking at the wrong moment is costly in both directions: a premature ``done'' sends the person on to a step they are not ready for, and a false alarm interrupts work that was going well---the moment of an interruption changes how much it costs~\cite{adamczyk2004if}, and false or repeated alerts make people less likely to act on the next one~\cite{dixon2007independence,ancker2017effects}.
The promise of video--language models is that these decisions can be read off the scene.

We find that one can score well on current benchmarks without reading the scene at all.
Consider a rule that never looks at the video: 25\,s into every step it raises an alarm, and at 30\,s it declares the step complete.
On Ego-MC-Bench~\cite{bhattacharyya2026streaming} this rule reaches a mistake F1 of 0.373, above every result reported for the benchmark (at most 0.27).
It confirms no step in time, but such rules trade alarms against confirmations along a whole curve, and with a learned policy on the same inputs they trace a frontier that every reported per-step result lies on or below (\cref{fig:frontier}; the revised scores of the benchmark paper in \suppref{sup:frontier}).

We call this curve the \emph{video-blind frontier} (\cref{sec:diagnosis}): the best trade-off between completion accuracy and mistake F1 reached by timing rules and a learned policy that see the elapsed time, the instruction and training-set statistics, but never the frames.
It contains all twelve zero-shot models that we evaluate on the Qualcomm Interactive Cooking dataset (QICD)~\cite{bhattacharyya2025livemamba}, from 2B models to Qwen3.8-Max, and a frontier of the same kind contains the speaking-time results published on GuideMe~\cite{liu2026guideme}.
The clock does not score because elapsed time predicts a single check---its AUROC is 0.52---but because the protocol accepts the first confirmation within a tolerance window, annotations cluster at step boundaries, and the typical duration of each instruction can be learned from the training split.
Models end up inside the frontier because the protocol asks every 5\,s and one alarm voids a step: in 83\% of the steps in which zero-shot Qwen3-VL-2B raises a false alarm, it does so at the very first check, it and Qwen3.8-Max lose more than half of the mistake-free steps to false alarms, and plainly fine-tuned models confirm steps before they end.

Both failures happen at moments when time has passed and the evidence has not yet arrived.
\emph{Evidence-Contrastive Training} (ECT, \cref{sec:method}) puts these moments into training.
It fine-tunes the model on the protocol's own checks and pairs positives with negatives from another step of the same recording at the same elapsed time (\cref{fig:teaser}).
The pair shares the kitchen, the person and the elapsed time, so elapsed time cannot separate its two sides.

The recipe is simple; its effect is not.
A 2B model trained with ECT on 209 videos, in about 15 GPU-hours and with neither synthetic data nor a decision threshold, lies beyond the frontier on both benchmarks in all four seeds, and does so on held-out halves of the test set with probability 0.87--0.97 on QICD and 0.99--1.00 on Ego-MC-Bench (\cref{sec:results}).
On Ego-MC-Bench, on which it is never trained, it confirms 59.0\% of the mistake-free steps in time, against 30.4\% for the same backbone fine-tuned on about 25k synthetic counterfactuals.
Its decisions follow the video beyond what timing explains (\cref{sec:evidence}).
At equal elapsed time it ranks a finished step above an unfinished one in 90--92\% of test pairs, where elapsed time alone scores 50\% and the instruction's typical duration 65\%.
Rescaling the time stamps written between the frames leaves it essentially unchanged, but costs the models trained without paired negatives, and a model trained on black frames, much of their separation.
Swapping in another recording, of the same recipe where one exists, drops its completion accuracy from 41.2 to 13.4.
The paired negatives work by cutting the share of checks that raise a false alarm from 18\% to 7\%, which accounts for most of the gain in completion accuracy.

\begin{figure*}[t]
  \centering
  \includegraphics[width=\linewidth]{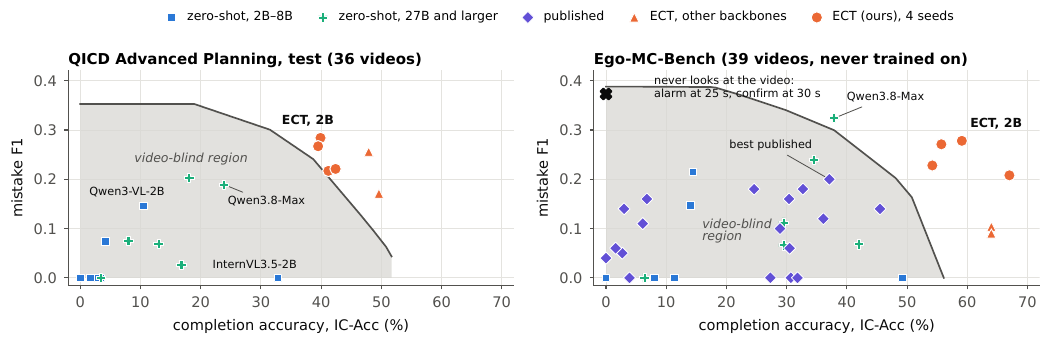}
  \caption{\textbf{The video-blind frontier contains every published Ego-MC-Bench result and every zero-shot model we run on QICD; ECT lies beyond it on both benchmarks in all four seeds.}
  The gray region is reached by policies that use only the elapsed time, the instruction and training-set statistics: timing rules and a learned policy, both tuned on the test set itself.
  Points are scores under the per-step protocol; published points are the 18 per-step results of Ego-MC-Bench v2, and its v3 results also lie inside. On Ego-MC-Bench a single rule (black cross) has higher mistake F1 than every published result, at zero completion accuracy; Qwen3.8-Max is the only zero-shot model beyond the frontier there.}
  \label{fig:frontier}
\end{figure*}

\noindent Our contributions are:
\begin{itemize}
  \item \textbf{A diagnosis and a tool.} We show that results on two guidance benchmarks can be matched without the video, formalize this as the video-blind frontier, explain why the clock scores, and recommend reporting this margin.
  \item \textbf{A simple method.} ECT supervises the protocol's own checks and pairs positives with negatives from the same recording at the same elapsed time, with no synthetic data and no decision threshold.
  \item \textbf{Evidence.} A 2B model trained with ECT lies beyond the frontier on both benchmarks. Time-matched pairs, rescaled time stamps and video swaps show that its decisions follow the video beyond what timing explains, and a replay of its decisions shows that paired negatives work mainly by cutting false alarms at each check.
\end{itemize}

\section{Related Work}
\label{sec:related}

\paragraph{Interactive guidance.}
Building on large egocentric video datasets~\cite{damen2018scaling,grauman2022ego4d}, HoloAssist~\cite{wang2023holoassist} recorded instructors guiding people through manual tasks and observed that instructors step in at once for critical mistakes.
QICD~\cite{bhattacharyya2025livemamba}, built on CaptainCook4D~\cite{peddi2024captaincook4d}, turned live cooking guidance into a benchmark that scores completion feedback and mistake feedback together; Ego-MC-Bench~\cite{bhattacharyya2026streaming} added recordings in which a human coach gives feedback in real time, and a model fine-tuned on about 25k synthetic counterfactuals; GuideMe~\cite{liu2026guideme} extends closed-loop guidance to several domains.
Offline, expert feedback can be generated or retrieved for a recorded attempt~\cite{ashutosh2025expertaf,ashutosh2026sportskills}.
We evaluate on the first two under their official protocol and re-read their published results against policies that do not see the video.

\paragraph{When to speak.}
Streaming video--language models process video as it arrives~\cite{chen2025livecc} and decide at every frame or chunk whether to respond~\cite{chen2024videollmonline,qian2025dispider}; benchmarks now score the timing of proactive responses~\cite{wang2025proactivevideoqa,zhang2025eyes}, and \citet{chatterjee2026dontpause} show that protocols that pause to ask hide the cost of late and missing responses.
In several learned systems the decision is governed by a parameter that sets how often the model speaks: a threshold on a silence token~\cite{chen2024videollmonline,zhang2025proactive,li2025lionfs}, a perplexity gate~\cite{yang2025livestar}, an activation threshold~\cite{wang2025streambridge}, or thresholds on dedicated heads and reward weights~\cite{wang2025videollm,wang2026mmduet2,yan2026proactvl}.
Others decide with a learned gate~\cite{ding2025streammind}, a trained trigger model~\cite{zheng2026emgarde} or a planning agent~\cite{yang2026anticipatory}.
Recent assistants tie speaking to the visual state of the task, through an explicit plan~\cite{kundu2026plan}, memory~\cite{gong2026vinci2} or learned speak decisions~\cite{kim2026egovoice,xu2026pro2assist}, and streaming question answering does so with completion checks~\cite{guo2026pace} or a readiness head trained with paired moments~\cite{azad2026streamready}; ProMemAssist~\cite{pu2025promemassist} instead times its messages from a model of the user's working memory.
We share that goal; we add no separate decision module, but change the training distribution so that elapsed time cannot explain the labels, and test whether the resulting timing depends on the video.

\paragraph{Shortcuts and complementary pairs.}
Balanced pairs and changed answer priors exposed language shortcuts in visual question answering~\cite{goyal2017making,agrawal2018dont}, and minimal pairs probe compositionality~\cite{thrush2022winoground} and visual grounding~\cite{tong2024eyes}; a single well-chosen frame answers much of video question answering~\cite{buch2022revisiting}, many multimodal benchmark items can be answered without the image~\cite{chen2024mmstar}, and time-reversed negatives teach models temporal order~\cite{bagad2023test}.
These works ask whether the \emph{answer} can be obtained without the intended visual input.
We ask whether the \emph{moment to speak} can, and because the answer depends on two coupled metrics we give a frontier rather than a single baseline.
\citet{umapathi2026ambient} notes that degenerate speaking policies score well on macro-F1.
The closest precedent is PROVIA~\cite{wen2026provia}, which requires online mistake detectors to beat perception-free controls built from the step index, the elapsed time and priors; we carry this idea over to interactive guidance.
Concurrently, \citet{benami2026have} show with a clock baseline that elapsed time is a shortcut for whether enough evidence has been seen in streaming question answering when negatives are sampled only before the evidence.
We study the sequential protocol of guidance benchmarks, intervene on the video at the level of the model's decisions, and train a model that moves beyond the frontier.
Our pairs hold the recording and the elapsed time fixed, so the cue they remove is the clock, and we use them both to train the model and to test it.

\paragraph{Procedural mistake detection.}
A separate line of work builds datasets of mistakes in recorded procedures~\cite{sener2022assembly101,peddi2024captaincook4d,lee2024error,haneji2025egoops} and detects or attributes them, offline~\cite{huang2025modeling,guo2026procedural,li2026mistake,ozsoy2026unreasonable} or online~\cite{flaborea2024prego,seminara2024differentiable}, surveyed by \citet{li2025building}.
\citet{spurio2026posttraining} post-train a vision--language model on pairs with the same step description and opposite evidence, for offline detection.
These methods score segments or actions; none is evaluated on confirming completion and flagging mistakes together under a given instruction, where every alarm is weighed against a confirmation.

\section{Do the benchmarks reward seeing? The video-blind frontier}
\label{sec:diagnosis}

On the two benchmarks we study, most reported scores can be reached without looking at the video.

\paragraph{Protocol and metrics.}
We use the official per-step protocol of QICD~\cite{bhattacharyya2025livemamba} and Ego-MC-Bench~\cite{bhattacharyya2026streaming}.
On QICD we use the Advanced Planning set, where the two decisions constrain each other most: its test split contains only recordings that deviate from the plan, and half of its steps contain a mistake.
The model is given the current step's instruction, and the input window is anchored at that step's annotated start. Only a model confirmation advances the instruction; if a step is never confirmed, later steps are not reached (\suppref{sup:protocol}).
Every 5\,s it receives the frames observed since the step began (at most 240) and is asked, with the official prompts and greedy decoding, whether the step is complete and, if not, whether a mistake has occurred; an answer is positive under the official parsing rule (it contains ``yes'').
We call each such query a \emph{check}.
In this protocol, ``zero-shot Qwen3-VL-2B'' denotes the stronger of its two variants, with the memory prompt of \citet{bhattacharyya2026streaming}, unless stated otherwise.
\emph{Completion accuracy} (IC-Acc) is the share of mistake-free steps whose completion is confirmed within 15\,s of the annotated step end; any mistake feedback issued during such a step voids it.
\emph{Mistake F1} matches mistake feedback to annotated mistakes with the same $\pm$15\,s tolerance.
The two metrics are coupled: every alarm can void a step that would otherwise count, and a model that never raises an alarm protects its completion accuracy at a mistake F1 of zero.
Neither number can be read alone.

\begin{table*}[t]
  \centering
  \small
  \setlength{\tabcolsep}{5pt}
  \resizebox{\textwidth}{!}{%
  \begin{tabular}{@{}llcccccc@{}}
    \toprule
    & & \multicolumn{3}{c}{QICD (Advanced Planning)} & \multicolumn{3}{c}{Ego-MC-Bench} \\
    \cmidrule(lr){3-5}\cmidrule(l){6-8}
    & Model & IC-Acc $\uparrow$ & F1 $\uparrow$ & $\Delta$ & IC-Acc $\uparrow$ & F1 $\uparrow$ & $\Delta$ \\
    \midrule
    \multirow{3}{*}{\rotatebox{90}{\scriptsize Blind}}
      & Rule with the best F1 (alarm at 25\,s, confirm at 30\,s) & 0.0 & 0.316 & -- & 0.0 & 0.373 & -- \\
      & Rule with the best IC-Acc & 46.6 & 0 & -- & 56.1 & 0 & -- \\
      & ECT trained and tested on black frames (2 seeds)$^{\flat}$ & 14.1 & 0 & \textcolor{black!62}{$-0.353$} & -- & -- & -- \\
    \midrule
    \multirow{7}{*}{\rotatebox{90}{\scriptsize Zero-shot}}
      & Qwen3-VL-2B & 4.2 & 0.074 & \textcolor{black!62}{$-0.279$} & 14.0 & 0.147 & \textcolor{black!62}{$-0.240$} \\
      & Qwen3-VL-2B + memory prompt & 10.5 & 0.146 & \textcolor{black!62}{$-0.207$} & 14.4 & 0.215 & \textcolor{black!62}{$-0.172$} \\
      & InternVL3.5-2B & 32.8 & 0 & \textcolor{black!62}{$-0.290$} & 49.2 & 0 & \textcolor{black!62}{$-0.187$} \\
      & LLaVA-OneVision-7B & 0.0 & 0 & \textcolor{black!62}{$-0.353$} & 0.0 & 0 & \textcolor{black!62}{$-0.388$} \\
      & Qwen2.5-VL-7B & 1.7 & 0 & \textcolor{black!62}{$-0.353$} & 8.0 & 0 & \textcolor{black!62}{$-0.388$} \\
      & Qwen3.5-397B & 18.1 & 0.203 & \textcolor{black!62}{$-0.150$} & 34.5 & 0.239 & \textcolor{black!62}{$-0.078$} \\
      & Qwen3.8-Max & 23.9 & 0.189 & \textcolor{black!62}{$-0.143$} & 37.9 & 0.324 & $+0.025$ \\
    \midrule
    \multirow{5}{*}{\rotatebox{90}{\scriptsize Published}}
      & Gemini-3-Flash, zero-shot~\cite{bhattacharyya2026streaming} & \multicolumn{3}{c}{\multirow{5}{*}{\parbox{5.4cm}{\centering\itshape\color{black!60} none published under this protocol\\ streaming protocol, LiveMamba~\cite{bhattacharyya2025livemamba}:\\ IC-Acc 12.6, F1 0.19}}} & 32.7 & 0.18 & \textcolor{black!62}{$-0.146$} \\
      & ProAssist~\cite{zhang2025proactive}, as reported in~\cite{bhattacharyya2026streaming} & \multicolumn{3}{c}{} & 3.0 & 0.14 & \textcolor{black!62}{$-0.248$} \\
      & Qwen3.5-2B, fine-tuned on QICD~\cite{bhattacharyya2026streaming} & \multicolumn{3}{c}{} & 28.9 & 0.10 & \textcolor{black!62}{$-0.244$} \\
      & Qwen3-VL-2B, fine-tuned on Ego-CoMist+~\cite{bhattacharyya2026streaming} & \multicolumn{3}{c}{} & 30.4 & 0.16 & \textcolor{black!62}{$-0.178$} \\
      & Qwen3.5-2B, fine-tuned on Ego-CoMist+~\cite{bhattacharyya2026streaming} & \multicolumn{3}{c}{} & 37.1 & 0.20 & \textcolor{black!62}{$-0.104$} \\
    \midrule
    \rowcolor{ectorange!16}
    \multirow{6}{*}{\rotatebox{90}{\scriptsize ECT}}
      & \textbf{Qwen3-VL-2B} (4 seeds) & $\mathbf{40.8}\pm1.3$ & $\mathbf{0.247}\pm0.033$ & $+0.01$ to $+0.06$ & $\mathbf{59.0}\pm5.7$ & $\mathbf{0.246}\pm0.034$ & $+0.17$ to $+0.28^{\dagger}$ \\
      & \quad same four models, thresholded decoding & $42.2\pm1.3$ & $0.293\pm0.035$ & $+0.08$ to $+0.12$ & $52.5\pm3.1$ & $0.242\pm0.041$ & $+0.01$ to $+0.21$ \\
      & \quad same fine-tuning without paired negatives (2 seeds) & 26.5 & 0.256 & \textcolor{black!62}{$-0.08$ to $-0.05$} & 48.3 & 0.337 & $+0.14$ \\
      & Qwen3-VL-4B & 40.3 & 0.227 & $+0.009$ & 61.7 & 0.175 & $+0.175^{\dagger}$ \\
      & InternVL3.5-2B & 47.9 & 0.256 & $+0.149$ & 64.0 & 0.104 & $+0.104^{\dagger}$ \\
      & LLaVA-OneVision-7B & 49.6 & 0.171 & $+0.090$ & 64.0 & 0.090 & $+0.090^{\dagger}$ \\
    \bottomrule
  \end{tabular}}
  \caption{\textbf{Who is beyond the video-blind frontier.}
  IC-Acc: completion accuracy (\%); F1: mistake F1; $\Delta$: mistake F1 minus the best video-blind F1 at the same or a higher IC-Acc, over the rules and the learned blind policy of \cref{sec:diagnosis} (negative, in gray: inside the frontier).
  $^{\dagger}$IC-Acc exceeds that of every blind policy, where the frontier is zero (two of the four seeds on Ego-MC-Bench).
  All rows except ``Published'' are our runs under the official per-step protocol; ECT is trained on QICD only.
  Rows with 4 or 2 seeds are means ($\pm$ standard deviation over the four seeds); every other row is a single run.
  Shaded: greedy decoding, our main setting; thresholded decoding uses two thresholds fixed on the validation split, and the row without negatives is that of \cref{tab:ablation}.
  Published rows are the scores reported in v2 of Ego-MC-Bench~\cite{bhattacharyya2026streaming} on 40 sessions (39 are available to us); the fine-tuned ones decide with a per-frame action head. The scores of its v3 also lie inside the frontier (\suppref{sup:frontier}).
  $^{\flat}$QICD only (\suppref{sup:results}).
  Model settings are in \suppref{sup:protocol}.}
  \label{tab:main}
\end{table*}

\paragraph{The video-blind frontier.}
A \emph{video-blind policy} decides when to confirm and when to raise an alarm from the time elapsed since the step began, the step index, the instruction text and statistics of the training split, and never from the frames.
We consider three families of rules: fixed delays (confirm completion $A$ seconds into every step, raise one alarm at $B$ seconds); confirmation at a multiple of the typical duration of the same instruction in the training split; and alarms only for instructions that often go wrong in the training split.
We add a learned policy on the same inputs: gradient-boosted trees fit on the QICD training and validation splits predict each step's duration and mistake probability, and a grid of policies confirms and alarms at multiples of the predicted duration.
We run every rule, per-step mixtures of rules and every learned policy through the official scorer, and define the \emph{video-blind frontier} as the upper convex hull of the resulting (IC-Acc, F1) points (\cref{fig:frontier}).
The \emph{margin} $\Delta$ of a result is its mistake F1 minus the frontier's F1 at the same or a higher IC-Acc; a negative margin means that a blind policy does at least as well on both metrics.
To favor the blind policies, we tune them on the test set itself.\footnote{Every vertex of the hull is a measured policy; between vertices we interpolate, which credits blind policies with trade-offs we did not run. Above the highest IC-Acc of any blind policy (51.7 on QICD, 56.1 on Ego-MC-Bench) the frontier is zero.}

\begin{figure*}[t]
  \centering
  \includegraphics[width=\linewidth]{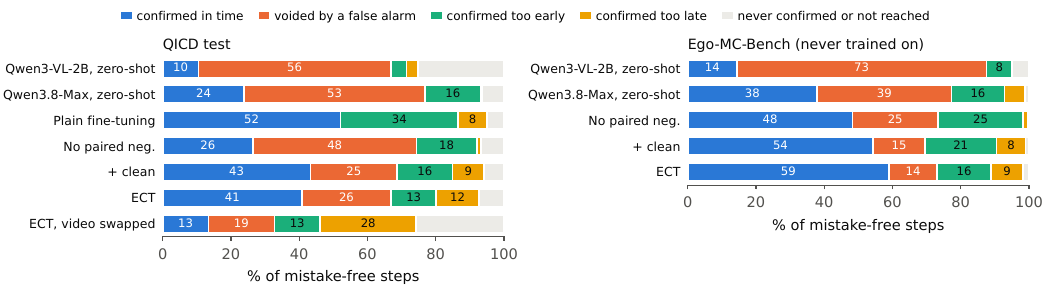}
  \caption{\textbf{Paired negatives reduce the steps lost to false alarms; swapping the video delays confirmation.}
  Outcome of every mistake-free step under official scoring (means over seeds; the swapped row is seed 0; grey: never confirmed or not reached).
  Zero-shot models lose steps to false alarms and plain fine-tuning to early confirmation. ECT confirms more steps in time than the variant without paired negatives on both benchmarks. In the QICD swap (last row, left), late and missing confirmations are more common.}
  \label{fig:decomp}
\end{figure*}

\paragraph{Comparing reported results with the frontier.}
A single rule---raise an alarm 25\,s into every step and confirm completion at 30\,s---reaches a mistake F1 of 0.373 on Ego-MC-Bench, above every reported result (at most 0.27); with its two delays chosen on the QICD validation split instead, it still reaches 0.361.
This rule confirms no step in time, which is why the comparison is made along the frontier.
None of the 18 per-step results in Tables~2 and~3 of Ego-MC-Bench~\cite{bhattacharyya2026streaming} lies beyond the frontier, in either version of that paper (\suppref{sup:frontier}); the result with the highest mistake F1 (33.0 / 0.27, fine-tuned Qwen3.5-2B in v3) has $\Delta=-0.054$.
The pattern is not specific to these benchmarks: on the speaking-time metrics of GuideMe~\cite{liu2026guideme}, all 13 results of the main table lie inside or on a frontier built from blind timing rules, and never speaking matches or exceeds 11 of them (\suppref{sup:frontier}).

\paragraph{Evaluating zero-shot models.}
Under the same protocol, all twelve zero-shot models that we run on QICD lie inside the frontier (\cref{fig:frontier}, \cref{tab:main}): six models with 2B to 8B parameters from three families, four sizes of Qwen3.5 from 27B to 397B, and two Qwen3.8 models.
Larger models are not reliably better here (the Qwen3.5 and Qwen3.8 models are given 16 frames per check), and most of the small models never raise an alarm.
On Ego-MC-Bench only Qwen3.8-Max crosses the frontier (37.9 / 0.324, $\Delta=+0.025$).

\paragraph{Why the clock scores.}
The reason is not that elapsed time predicts the answer to an individual check: over the checks of the QICD test set its area under the ROC curve (AUROC) is about 0.52 for either question.
The clock scores because the protocol is sequential.
The first confirmation ends a step and is accepted within $\pm$15\,s, so a rule only has to choose when to speak first, not to be right at every check.
And the annotations sit where a clock can find them: 91\% of the mistake feedback in QICD is annotated exactly at the end of the step.
Where recipes repeat, as in QICD, the training split adds the typical duration of each instruction: relative to it, elapsed time separates completions at an AUROC of 0.61, a signal that fine-tuning on the protocol's checks can pick up. On Ego-MC-Bench only 2 of 562 instructions occur in the training split (\suppref{sup:frontier}).

\paragraph{Two ways to end up inside.}
Decomposing each model's decisions on mistake-free steps by their outcome under the official scorer, in the spirit of TIDE~\cite{bolya2020tide}, shows two failure modes (\cref{fig:decomp}).
Some models \emph{raise alarms without evidence}: a false alarm voids 56.3\% of the mistake-free steps for zero-shot Qwen3-VL-2B and 52.9\% for Qwen3.8-Max. Because the protocol asks every 5\,s and a single alarm voids the step, a modest rate per check becomes a large loss per step; the zero-shot 2B model raises its first false alarm at the very first check in 83\% of the steps it voids.
Others \emph{confirm too early}: after plain fine-tuning on the QICD training split, Qwen3-VL-2B never raises an alarm and confirms 34.5\% of the mistake-free steps before they end.
Both failures happen at moments when time has passed but the evidence has not arrived; \cref{sec:method} puts exactly these moments into training.

\paragraph{Recommendation.}
Because the two metrics are coupled, either one alone can be raised by a rule that does not look at the video.
We recommend reporting the margin $\Delta$ over the video-blind frontier alongside results on these benchmarks; \suppref{sup:frontier} lists its vertices.

\section{Evidence-Contrastive Training}
\label{sec:method}

Both failures in \cref{sec:diagnosis} occur at moments when time has passed and the evidence has not arrived.
Evidence-Contrastive Training (ECT) puts these moments into training.
It has two ingredients: each training example is a check of the benchmark protocol, and positives are paired with negatives at the same elapsed time.

\paragraph{Situations.}
Every training example is one check of the protocol, which we call a \emph{situation}: the frames since the step began, the official prompt, and the answer.
From the QICD annotations we take \emph{mistake} situations at the annotated feedback time and 5 and 10\,s later, answered ``YES'' followed by the feedback, and \emph{success} situations at the step end and 5\,s later, answered ``YES''.
Checks at which nothing should be said are answered ``NO''; these include checks placed before an annotated mistake, outside its tolerance window.
During training, situations are sampled by kind with fixed weights (\suppref{sup:training}).

\paragraph{Paired negatives.}
For a positive at elapsed time $E$, ECT looks for a negative from another step of the same recording at the same $E$ (\cref{fig:teaser} shows one such pair); we call these \emph{paired negatives}.
A \emph{notdone} negative pairs a success situation with a step that still has at least 20\,s to run, and answers ``NO'' to the completion question; it targets early confirmation.
A \emph{clean} negative pairs a mistake situation with a mistake-free step that still has at least 15\,s to run, and answers ``NO'' to the mistake question; it targets alarms without evidence.
Not every positive has such a partner in its recording: we obtain 4,068 notdone negatives for 4,563 success situations and 1,159 clean negatives for 1,938 mistake situations.

\paragraph{What the pair controls.}
Because both sides come from one recording, the kitchen, the person and the camera are shared; because the elapsed time is equal, so are the window length and the clock reading.
Within a pair, any policy that reads only the elapsed time or the window length must give both sides the same answer; the instruction can still differ, and with it the step's typical duration.
Pooled over all positives and their paired negatives, elapsed time separates the labels with an AUROC of 0.541 for the completion question and 0.606 for the mistake question, since not every positive has a partner.

\paragraph{Training and inference.}
We fine-tune Qwen3-VL-2B~\cite{bai2025qwen3vl} with rank-64 LoRA~\cite{hu2022lora} for 2,400 steps, about one epoch over the roughly 18,400 situations built from the 209 training videos, with cross-entropy on the answer tokens; this takes about 15 GPU-hours on a single A100-40G.
At test time the model is run exactly as in \cref{sec:diagnosis}: the official prompts and greedy decoding.
ECT uses no decision threshold and no synthetic data.

\section{Results}
\label{sec:results}

\paragraph{Setup.}
ECT is trained on the 209 training videos of QICD Advanced Planning and evaluated on its 36 test videos and, without any adaptation, on the 39 available videos of Ego-MC-Bench.
The two benchmarks are complementary: in QICD mistake feedback is almost always annotated at the end of a step, whereas in Ego-MC-Bench a human coach gives it while the step is under way (4\% at the step end).
Each test configuration is run once.
We train four seeds, report the mean and standard deviation over seeds, and test differences with a paired bootstrap over videos~\cite{efron1979bootstrap} and Holm correction~\cite{holm1979simple}.

\paragraph{ECT lies beyond the frontier.}
All four seeds lie beyond the video-blind frontier on both benchmarks (\cref{tab:main}, \cref{fig:frontier}).
On QICD, ECT reaches $40.8\pm1.3$ IC-Acc and $0.247\pm0.033$ mistake F1, with margins from $+0.012$ to $+0.061$; on Ego-MC-Bench, on which it is never trained, $59.0\pm5.7$ and $0.246\pm0.034$, with margins from $+0.17$ to $+0.28$.
Two of the Ego-MC-Bench seeds exceed the completion accuracy of every blind policy, where the frontier is zero; the other two have margins of $+0.169$ and $+0.259$ against a non-zero frontier.
The pre-registered variant with decision thresholds chosen on validation data is also beyond the frontier in every seed (\cref{tab:main}).
When the blind policies are tuned on one half of the test videos and the margin is measured on the other, a seed lies beyond the frontier in 87--97\% of 500 splits on QICD and 99--100\% on Ego-MC-Bench (\suppref{sup:frontier}).

\paragraph{Against published fine-tuned models.}
The reported Qwen3-VL-2B fine-tuned on about 25k synthetic counterfactual pairs mixed with QICD and Goal-Step reaches 30.4 / 0.16 (0.20 in the revised version), and a reported Qwen3.5-2B trained on QICD alone reaches 28.9 / 0.10~\cite{bhattacharyya2026streaming}.
ECT uses 209 QICD videos and 2,400 steps on one A100, versus about 100k iterations on eight H100 GPUs for the reported same-backbone model. ECT reaches 59.0 / 0.246 on 39 released sessions; the published scores above use 40 sessions.
Our fine-tuning without paired negatives already reaches 48.3 / 0.337; adding the negatives raises completion accuracy but lowers mistake F1 (\cref{tab:main}).
QICD Advanced Planning has no published result under the per-step protocol, so on that benchmark we compare with blind policies and zero-shot models.

\paragraph{Against zero-shot models.}
Against the same model zero-shot, ECT raises completion accuracy from 10.5 to 40.8 on QICD and from 14.4 to 59.0 on Ego-MC-Bench (significant for every seed tested); the gain in mistake F1 is significant for two of the four seeds on QICD and not on Ego-MC-Bench (0.246 against 0.215), where 0.43 of its alarms are correct, against 0.20 zero-shot and 0.32 for Qwen3.8-Max.
Qwen3.8-Max crosses the frontier on one benchmark, by $+0.025$; the 2B model with ECT crosses on both and reaches 40.8\% IC-Acc on QICD, against 23.9\% for Qwen3.8-Max.

\paragraph{Other backbones and sizes.}
The same recipe moves InternVL3.5-2B~\cite{wang2025internvl35} (32.8 / 0 zero-shot to 47.9 / 0.256) and LLaVA-OneVision-7B~\cite{li2025llavaonevision} (0 / 0 to 49.6 / 0.171) beyond the frontier on QICD.
On Ego-MC-Bench both reach 64.0 IC-Acc, above every blind policy (56.1), but their mistake recall is only 0.05--0.07, so across backbones we claim only the gain in completion accuracy.
The same recipe also puts one Qwen3-VL-4B seed beyond the frontier on both benchmarks (40.3 / 0.227 and 61.7 / 0.175), without a consistent improvement over the 2B model across both metrics.
Results with smaller training subsets are in \suppref{sup:results}.

\paragraph{Streaming protocol.}
Our diagnosis concerns the per-step protocol: in the streaming setting of the reported QICD Advanced Planning results, without the annotated step-start anchors, blind rules are weak because their errors accumulate. The four seeds reach an IC-Acc of 16.4--22.3 (\suppref{sup:results}) against 12.6 for LiveMamba~\cite{bhattacharyya2025livemamba}, whose mistake F1 (0.19) is higher than ours (0.13--0.18) and whose training data differ.
On the Main set of QICD, where two published systems lie beyond the frontier~\cite{bhattacharyya2025livemamba,var2026aicoach}, all four greedy ECT seeds do so without tuning on that set, with margins of $+0.072$ to $+0.143$ (\suppref{sup:results}).

\section{Does the model wait for evidence?}
\label{sec:evidence}

\begingroup
\widowpenalty=10000
\clubpenalty=10000
\brokenpenalty=10000

Where elapsed time is held fixed, ECT decides by the video well beyond what the instruction's typical duration explains, and it does not respond to the time stamps between the frames. Paired negatives get it there mainly by cutting false alarms at each check.

\paragraph{Swapping the video.}
We replace each test video with another recording, of the same recipe wherever one exists, keeping the instructions, the step boundaries and the timing.
A model that follows the clock would not notice.
For seed~0 on the test set, completion accuracy falls from 41.2 to 13.4 and mistake F1 from 0.217 to 0.094. On the validation split, completion accuracy falls from 47.2 to 14.7 (\suppref{sup:perturb}).
On that validation split, shuffling the frames costs 10 completion points, and repeating only the last frame reduces completion accuracy to 0.7.
On black frames ECT says nothing at all, while the same fine-tuning without paired negatives still raises 109 false alarms.
The way it fails changes too (\cref{fig:decomp}, last row): the share of steps voided by a false alarm falls from 27\% to 19\%, while late confirmations rise from 12\% to 28\% and a further 21\% of the steps are never confirmed or never reached.
Deprived of evidence, the model waits (\cref{fig:swap}).

\begin{figure}[t]
  \centering
  \includegraphics[width=\linewidth]{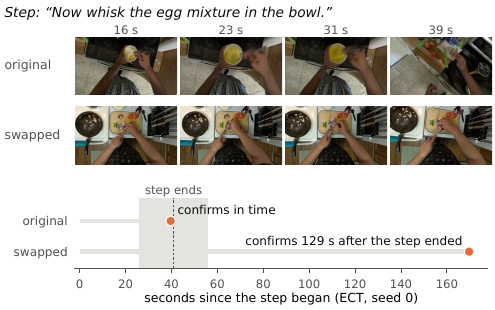}
  \caption{\textbf{Without the matching video, ECT waits.}
  A mistake-free QICD test step under the original recording (top) and under another recording of the same recipe at the same frame indices (bottom); instruction and clock are unchanged.
  ECT confirms the step on time in the original and 129\,s late in the swap, without raising an alarm. Gray: the $\pm$15\,s tolerance around the step end.
  The step is one of the seven nearest the median duration among the 28 test steps that seed~0 confirms on time and, after the swap, neither flags nor confirms within tolerance (\suppref{sup:perturb}).}
  \label{fig:swap}
\end{figure}

\begin{table*}[t]
  \centering
  \small
  \setlength{\tabcolsep}{6.5pt}
  \begin{tabular}{@{}lccccccccccc@{}}
    \toprule
    & & \multicolumn{3}{c}{QICD} & \multicolumn{3}{c}{Ego-MC-Bench} & \multicolumn{3}{c}{QICD, mechanism} \\
    \cmidrule(lr){3-5}\cmidrule(lr){6-8}\cmidrule(l){9-11}
    Training & $n$ & IC-Acc $\uparrow$ & F1 $\uparrow$ & $\Delta$ & IC-Acc $\uparrow$ & F1 $\uparrow$ & $\Delta$ & Void $\downarrow$ & Early $\downarrow$ & Pair $\uparrow$ \\
    \midrule
    Plain fine-tuning$^{\ddagger}$ & 1 & 52.1 & 0.000 & $0.000$ & 60.2 & 0.000 & $0.000$ & 0 & 35 & -- \\
    No neg., re-balanced sampling & 1 & 27.7 & 0.321 & $+0.005$ & 47.7 & 0.318 & $+0.112$ & 40 & 22 & 0.67 \\
    \midrule
    No paired neg. & 2 & 26.5 & 0.256 & $-0.066$ & 48.3 & 0.337 & $+0.139$ & 48 & 18 & 0.65 \\
    + notdone$^{*}$ & 4 & 34.0 & 0.297 & $+0.021$ & 54.0 & 0.283 & $+0.207$ & 40 & 15 & 0.72 \\
    + clean & 2 & 43.3 & 0.286 & $+0.112$ & 54.4 & 0.222 & $+0.168$ & 25 & 16 & 0.74 \\
    \rowcolor{ectorange!16}
    \textbf{ECT} (both) & 4 & 40.8 & 0.247 & $+0.036$ & 59.0 & 0.246 & $+0.229$ & 26 & 13 & 0.76 \\
    Random-moment neg.$^{\S}$ & 2 & 38.7 & 0.281 & $+0.048$ & -- & -- & -- & 31 & 13 & 0.74 \\
    \bottomrule
  \end{tabular}
  \captionsetup{justification=raggedright}
  \caption[What each ingredient changes]{\textbf{What each ingredient changes.}
  All rows fine-tune Qwen3-VL-2B and use greedy decoding; means over $n$ training seeds. $\Delta$: margin over the video-blind frontier. Void / Early: share (\%) of mistake-free steps voided by a false alarm / confirmed too early. Pair: accuracy on time-matched mistake pairs (0.50 for any policy of elapsed time alone).
  ECT is to be compared with ``No paired neg.'', which differs from it in the negatives alone.
  \par\smallskip\footnotesize
  $^{*}$Two seeds on Ego-MC-Bench. $^{\ddagger}$An earlier recipe that also differs in how situations are placed and sampled (\suppref{sup:training}); it never raises an alarm. $^{\S}$ECT with negatives drawn at random moments of the same steps; same-seed comparisons, including Ego-MC-Bench, are in \suppref{sup:results}.}
  \label{tab:ablation}
\end{table*}

\paragraph{Time-matched test pairs.}
A swap changes the video and can also change its alignment with the instruction.
To test whether elapsed time alone explains the decisions, we build 931 pairs from the QICD test videos as in ECT's negative sampling: two checks from the same recording, at the same elapsed time and with the same window length to within one frame in 928 pairs (the other three are cut short by the end of the video). They ask the same question with opposite answers, but the step instructions can differ.
We measure how often a model gives the higher probability of ``yes'' to the correct side; any policy of elapsed time alone scores exactly 0.5, and one that also uses the instruction's typical duration in the training split scores 0.65 on completion and 0.54 on mistake pairs, because the two steps of a pair usually differ.
Zero-shot Qwen3-VL-2B (a single check, so without the memory prompt) scores 0.66 on completion pairs and 0.59 on mistake pairs; ECT scores 0.90--0.92 and 0.75--0.77 over four seeds (\suppref{sup:pairs}).
The three models fine-tuned without paired negatives (two seeds and the re-balanced variant of \cref{tab:ablation}) reach 0.63--0.67 on mistake pairs: the negatives add $+0.104$ (95\% CI $[+0.044, +0.156]$, paired bootstrap over videos) and cut the rate of answering ``yes'' on the side without evidence from 23--32\% to 4--8\%.
The gain carries over to InternVL3.5-2B (0.72 against 0.67), and LLaVA-OneVision-7B with ECT reaches 0.75.
On completion pairs every fine-tuned variant scores 0.87--0.92: supervising the protocol's checks already teaches the model to see that a step is finished.

\paragraph{Time stamps.}
Qwen3-VL writes a time stamp between the frames of a window. On 411 fixed validation checks we doubled or halved every stamp, keeping the pixels, the number of frames and the prompt (\suppref{sup:perturb}).
ECT (seed 0) is unaffected: its AUROC is 0.945, 0.942 and 0.943 for completion and 0.841, 0.820 and 0.846 for mistakes with the original, doubled and halved stamps.
Without paired negatives, the same check separates mistakes from silence at 0.971 with the original stamps and 0.79--0.80 with rescaled ones, no longer above ECT; both rescalings raise its probability of ``yes'' on silent checks, so it reacts to unfamiliar stamps rather than reading their value. After plain fine-tuning the separation drops from 0.961 to 0.60--0.61.
The ECT recipe trained on black frames separates completions at 0.920 without any video, and 0.74--0.77 once the stamps are rescaled.
ECT still uses how long a step usually takes, as a coach would: its confirmation times follow the median training duration of the instruction, a cue its training pairs leave in place. At equal elapsed time, though, it scores 0.90--0.92 on the pairs where this cue reaches 0.65 and the black-frame model 0.62--0.64.

\paragraph{What the negatives change.}
Fine-tuning on the protocol's checks improves completion accuracy; the paired negatives change the balance between confirmations and alarms (\cref{tab:ablation}).
Fine-tuning without paired negatives already puts the model beyond the frontier on Ego-MC-Bench ($\Delta=+0.139$); on QICD it stays inside ($\Delta=-0.066$).
It gets there by raising alarms without evidence: 48\% of the mistake-free steps are voided, with 1.3--1.5 false alarms per mistake-free step (0.5--0.6 for ECT), and completion accuracy on QICD is 26.5.
Clean negatives halve the voided steps (25\%) and lift completion accuracy to 43.3, with the largest margin on QICD.
Re-balancing the answers by sampling, instead of adding negatives, keeps the completion accuracy low and the false alarms in place.

Notdone negatives make the confirmation follow the progress of the step: on the unseen steps of Ego-MC-Bench its correlation with the step's duration is 0.45--0.69 for the variants trained with them (0.68 with notdone negatives alone), 0.30--0.40 for those without and 0.10 for the zero-shot model, and fewer mistake-free steps are confirmed more than 15\,s early (37--55 against 62--82; \suppref{sup:results}).
Relative to no paired negatives, using both kinds reduces early confirmation from 18\% to 13\% on QICD and from 25\% to 16\% on Ego-MC-Bench, where IC-Acc rises from 48.3 to 59.0.

Negatives drawn at random moments of the same steps also put both tested seeds beyond the frontier on both benchmarks: within a recording the clock separates them from the positives about as poorly as matched ones (AUROC 0.555--0.561 and 0.607--0.620, against 0.541 and 0.606), and they shift the trade-off between the two metrics (\suppref{sup:results}).

\paragraph{Where the gain comes from.}
For ECT and the variant without paired negatives (seed 0 of each), we logged both probabilities at every check of the QICD test set and replayed the official protocol offline; at the thresholds of the pre-registered decoding the replay matches the official IC-Acc of both runs (42.0 and 31.1), counting the few steps that outlast the log as confirmed at the next check (\suppref{sup:decomp}).
Over the checks of mistake-free steps, 7.3\% raise a false alarm for ECT and 18.3\% for the variant, and 26.5\% and 40.8\% of these steps are voided.
Giving the variant the mistake probabilities of ECT, with its own completion probabilities, raises its IC-Acc from 31.1 to 40.8, close to the 42.0 of ECT (23.1 to 38.7 against 39.1 at thresholds of 0.5, the closest to greedy decoding): fewer false alarms per check account for most of the gain in completion accuracy.
The completion side improves too: with every alarm removed, ECT confirms more steps in time than the variant at every completion threshold from 0.1 to 0.6 (at best 61.3 against 54.6).

\paragraph{The operating point.}
The negatives trade mistake recall for confirmations: without them mistake F1 on Ego-MC-Bench is higher (0.34) at a completion accuracy 11 points lower, because the model raises alarms in about twice as many steps, at a precision of 0.33 against 0.43 (\suppref{sup:results}), and tripling the clean negatives moves QICD from 40.8 / 0.247 to 52.5 / 0.222.
How many negatives to use is therefore a choice of operating point. Given the same validation cap of 432 raw alarms, the two variants without clean negatives fall 7.1 and 5.9 completion points short of thresholded ECT on QICD (\suppref{sup:operating}).
\par
\endgroup

\Needspace{14\baselineskip}
\section{Limitations and Conclusion}
\label{sec:conclusion}

\paragraph{Limitations.}
Our study covers cooking, two benchmarks and their official protocol, and Ego-MC-Bench is evaluated by transfer.
A blind learner with more inputs than our definition allows can extend the frontier: one that also sees the durations of the earlier steps of the same recording, which the protocol reveals, brings the QICD margins of greedy ECT to $-0.028$--$+0.029$, while those on Ego-MC-Bench and the QICD Main set stay positive (\suppref{sup:frontier}).
ECT improves the timing of what the model says more than the mistakes it catches: its mistake recall stays low (0.18 on Ego-MC-Bench), it separates mistakes from mistake-free completions at a step end less well than fine-tuning without paired negatives (\suppref{sup:pairs}), and a detection is credited by its time alone. In a manual check of 25 detections, 9 named the annotated mistake, 3 named it in part, and 13 named a different one (\suppref{sup:content}).

\paragraph{Conclusion.}
Under their per-step protocol, two benchmarks for proactive guidance currently reward the clock, and the video-blind frontier measures by how much.
ECT moves a 2B model beyond it on both benchmarks without synthetic data or a decision threshold. At equal elapsed time its decisions are set by the video, it ignores the time stamps that a blind model relies on, and its paired negatives work by cutting false alarms at each check. We recommend reporting the margin over the frontier with every result on these benchmarks.

{\small
\let\originalbibsection\bibsection
\renewcommand{\bibsection}{\pdfbookmark[1]{\refname}{preprint.references}\originalbibsection\label{sec:references}}
\bibliographystyle{ieeenat_fullname}\bibliography{main}}

\begin{thebibliography}{64}
\providecommand{\natexlab}[1]{#1}
\providecommand{\url}[1]{\texttt{#1}}
\expandafter\ifx\csname urlstyle\endcsname\relax
  \providecommand{\doi}[1]{doi: #1}\else
  \providecommand{\doi}{doi: \begingroup \urlstyle{rm}\Url}\fi

\bibitem[Adamczyk and Bailey(2004)]{adamczyk2004if}
Piotr~D. Adamczyk and Brian~P. Bailey.
\newblock If not now, when?: The effects of interruption at different moments
  within task execution.
\newblock In \emph{Proceedings of the SIGCHI Conference on Human Factors in
  Computing Systems (CHI)}, pages 271--278. ACM, 2004.

\bibitem[Agrawal et~al.(2018)Agrawal, Batra, Parikh, and
  Kembhavi]{agrawal2018dont}
Aishwarya Agrawal, Dhruv Batra, Devi Parikh, and Aniruddha Kembhavi.
\newblock Don't just assume; look and answer: Overcoming priors for visual
  question answering.
\newblock In \emph{Proceedings of the IEEE Conference on Computer Vision and
  Pattern Recognition (CVPR)}, pages 4971--4980, 2018.

\bibitem[Ancker et~al.(2017)Ancker, Edwards, Nosal, Hauser, Mauer, Kaushal, and
  {the HITEC Investigators}]{ancker2017effects}
Jessica~S. Ancker, Alison Edwards, Sarah Nosal, Diane Hauser, Elizabeth Mauer,
  Rainu Kaushal, and {the HITEC Investigators}.
\newblock Effects of workload, work complexity, and repeated alerts on alert
  fatigue in a clinical decision support system.
\newblock \emph{BMC Medical Informatics and Decision Making}, 17\penalty0
  (1):\penalty0 36, 2017.

\bibitem[Ashutosh et~al.(2025)Ashutosh, Nagarajan, Pavlakos, Kitani, and
  Grauman]{ashutosh2025expertaf}
Kumar Ashutosh, Tushar Nagarajan, Georgios Pavlakos, Kris Kitani, and Kristen
  Grauman.
\newblock {ExpertAF}: Expert actionable feedback from video.
\newblock In \emph{Proceedings of the IEEE/CVF Conference on Computer Vision
  and Pattern Recognition (CVPR)}, pages 13582--13594, 2025.

\bibitem[Ashutosh et~al.(2026)Ashutosh, Wu, and
  Grauman]{ashutosh2026sportskills}
Kumar Ashutosh, Chi~Hsuan Wu, and Kristen Grauman.
\newblock {SportSkills}: Physical skill learning from sports instructional
  videos, 2026.
\newblock arXiv preprint arXiv:2603.25163.

\bibitem[Azad et~al.(2026)Azad, Vineet, and Rawat]{azad2026streamready}
Shehreen Azad, Vibhav Vineet, and Yogesh~S Rawat.
\newblock {StreamReady}: Learning what to answer and when in long streaming
  videos.
\newblock In \emph{Proceedings of the IEEE/CVF Conference on Computer Vision
  and Pattern Recognition (CVPR)}, pages 40494--40504, 2026.

\bibitem[Bagad et~al.(2023)Bagad, Tapaswi, and Snoek]{bagad2023test}
Piyush Bagad, Makarand Tapaswi, and Cees G.~M. Snoek.
\newblock Test of time: Instilling video-language models with a sense of time.
\newblock In \emph{Proceedings of the IEEE/CVF Conference on Computer Vision
  and Pattern Recognition (CVPR)}, pages 2503--2516, 2023.

\bibitem[Bai et~al.(2025)Bai, Cai, Chen, Chen, Chen, Cheng, Deng, Ding, Gao,
  Ge, Ge, Guo, Huang, Huang, Huang, Hui, Jiang, Li, Li, Li, Li, Lin, Lin, Liu,
  Liu, Liu, Liu, Liu, Liu, Lu, Luo, Lv, Men, Meng, Ren, Ren, Song, Sun, Tang,
  Tu, Wan, Wang, Wang, Wang, Wang, Xie, Xu, Xu, Xu, Yang, Yang, Yang, Yang, Yu,
  Zhang, Zhang, Zhang, Zheng, Zhong, Zhou, Zhou, Zhou, Zhu, and
  Zhu]{bai2025qwen3vl}
Shuai Bai, Yuxuan Cai, Ruizhe Chen, Keqin Chen, Xionghui Chen, Zesen Cheng,
  Lianghao Deng, Wei Ding, Chang Gao, Chunjiang Ge, Wenbin Ge, Zhifang Guo,
  Qidong Huang, Jie Huang, Fei Huang, Binyuan Hui, Shutong Jiang, Zhaohai Li,
  Mingsheng Li, Mei Li, Kaixin Li, Zicheng Lin, Junyang Lin, Xuejing Liu,
  Jiawei Liu, Chenglong Liu, Yang Liu, Dayiheng Liu, Shixuan Liu, Dunjie Lu,
  Ruilin Luo, Chenxu Lv, Rui Men, Lingchen Meng, Xuancheng Ren, Xingzhang Ren,
  Sibo Song, Yuchong Sun, Jun Tang, Jianhong Tu, Jianqiang Wan, Peng Wang,
  Pengfei Wang, Qiuyue Wang, Yuxuan Wang, Tianbao Xie, Yiheng Xu, Haiyang Xu,
  Jin Xu, Zhibo Yang, Mingkun Yang, Jianxin Yang, An Yang, Bowen Yu, Fei Zhang,
  Hang Zhang, Xi Zhang, Bo Zheng, Humen Zhong, Jingren Zhou, Fan Zhou, Jing
  Zhou, Yuanzhi Zhu, and Ke Zhu.
\newblock {Qwen3-VL} technical report, 2025.
\newblock arXiv preprint arXiv:2511.21631.

\bibitem[Ben-Ami et~al.(2026)Ben-Ami, Cohen, and Baskin]{benami2026have}
Dan Ben-Ami, Kobi Cohen, and Chaim Baskin.
\newblock Have {I} seen enough? frozen video-language models encode evidence
  readiness, 2026.
\newblock arXiv preprint arXiv:2610.08560.

\bibitem[Bhattacharyya et~al.(2025)Bhattacharyya, Xu, Haresh, Pourreza, Liu,
  Panchal, Madan, Sigal, and Memisevic]{bhattacharyya2025livemamba}
Apratim Bhattacharyya, Bicheng Xu, Sanjay Haresh, Reza Pourreza, Litian Liu,
  Sunny Panchal, Pulkit Madan, Leonid Sigal, and Roland Memisevic.
\newblock Can multi-modal {LLMs} provide live step-by-step task guidance?
\newblock In \emph{Advances in Neural Information Processing Systems}, pages
  22377--22410, 2025.

\bibitem[Bhattacharyya et~al.(2026)Bhattacharyya, Mahajan, Haresh, Pourreza,
  Liu, Yasarla, Garrepalli, and Memisevic]{bhattacharyya2026streaming}
Apratim Bhattacharyya, Shweta Mahajan, Sanjay Haresh, Reza Pourreza, Litian
  Liu, Rajeev Yasarla, Risheek Garrepalli, and Roland Memisevic.
\newblock Streaming interventions: Can video {LLMs} correct mistakes as they
  occur?
\newblock In \emph{Advances in Neural Information Processing Systems
  (Evaluations \& Datasets Track)}, 2026.
\newblock Benchmark scores compared against
  \href{https://arxiv.org/abs/2606.09547v2}{arXiv:2606.09547v2} and
  \href{https://arxiv.org/abs/2606.09547v3}{v3}.

\bibitem[Bolya et~al.(2020)Bolya, Foley, Hays, and Hoffman]{bolya2020tide}
Daniel Bolya, Sean Foley, James Hays, and Judy Hoffman.
\newblock {TIDE}: A general toolbox for identifying object detection errors.
\newblock In \emph{Proceedings of the European Conference on Computer Vision
  (ECCV)}, pages 558--573. Springer, 2020.

\bibitem[Buch et~al.(2022)Buch, Eyzaguirre, Gaidon, Wu, Fei-Fei, and
  Niebles]{buch2022revisiting}
Shyamal Buch, Crist{\'o}bal Eyzaguirre, Adrien Gaidon, Jiajun Wu, Li Fei-Fei,
  and Juan~Carlos Niebles.
\newblock Revisiting the ``video'' in video-language understanding.
\newblock In \emph{Proceedings of the IEEE/CVF Conference on Computer Vision
  and Pattern Recognition (CVPR)}, pages 2917--2927, 2022.

\bibitem[Chatterjee et~al.(2026)Chatterjee, Pang, Sener, Song, and
  Yao]{chatterjee2026dontpause}
Dibyadip Chatterjee, Zhanzhong Pang, Fadime Sener, Yale Song, and Angela Yao.
\newblock Don't pause! every prediction matters in a streaming video.
\newblock In \emph{Advances in Neural Information Processing Systems}, 2026.

\bibitem[Chen et~al.(2024{\natexlab{a}})Chen, Lv, Wu, Lin, Song, Gao, Liu, Gao,
  Mao, and Shou]{chen2024videollmonline}
Joya Chen, Zhaoyang Lv, Shiwei Wu, Kevin~Qinghong Lin, Chenan Song, Difei Gao,
  Jia-Wei Liu, Ziteng Gao, Dongxing Mao, and Mike~Zheng Shou.
\newblock {VideoLLM-online}: Online video large language model for streaming
  video.
\newblock In \emph{Proceedings of the IEEE/CVF Conference on Computer Vision
  and Pattern Recognition (CVPR)}, pages 18407--18418, 2024{\natexlab{a}}.

\bibitem[Chen et~al.(2025)Chen, Zeng, Lin, Li, Ma, and Shou]{chen2025livecc}
Joya Chen, Ziyun Zeng, Yiqi Lin, Wei Li, Zejun Ma, and Mike~Zheng Shou.
\newblock {LiveCC}: Learning video {LLM} with streaming speech transcription at
  scale.
\newblock In \emph{Proceedings of the IEEE/CVF Conference on Computer Vision
  and Pattern Recognition (CVPR)}, pages 29083--29095, 2025.

\bibitem[Chen et~al.(2024{\natexlab{b}})Chen, Li, Dong, Zhang, Zang, Chen,
  Duan, Wang, Qiao, Lin, and Zhao]{chen2024mmstar}
Lin Chen, Jinsong Li, Xiaoyi Dong, Pan Zhang, Yuhang Zang, Zehui Chen, Haodong
  Duan, Jiaqi Wang, Yu Qiao, Dahua Lin, and Feng Zhao.
\newblock Are we on the right way for evaluating large vision-language models?
\newblock In \emph{Advances in Neural Information Processing Systems}, pages
  27056--27087, 2024{\natexlab{b}}.

\bibitem[Damen et~al.(2018)Damen, Doughty, Farinella, Fidler, Furnari, Kazakos,
  Moltisanti, Munro, Perrett, Price, and Wray]{damen2018scaling}
Dima Damen, Hazel Doughty, Giovanni~Maria Farinella, Sanja Fidler, Antonino
  Furnari, Evangelos Kazakos, Davide Moltisanti, Jonathan Munro, Toby Perrett,
  Will Price, and Michael Wray.
\newblock Scaling egocentric vision: The {EPIC-KITCHENS} dataset.
\newblock In \emph{Proceedings of the European Conference on Computer Vision
  (ECCV)}, pages 753--771, 2018.

\bibitem[Ding et~al.(2025)Ding, Wu, Yang, Jiang, Zhang, Bai, Chen, and
  Cao]{ding2025streammind}
Xin Ding, Hao Wu, Yifan Yang, Shiqi Jiang, Qianxi Zhang, Donglin Bai, Zhibo
  Chen, and Ting Cao.
\newblock {StreamMind}: Unlocking full frame rate streaming video dialogue
  through event-gated cognition.
\newblock In \emph{Proceedings of the IEEE/CVF International Conference on
  Computer Vision (ICCV)}, pages 13448--13459, 2025.

\bibitem[Dixon et~al.(2007)Dixon, Wickens, and McCarley]{dixon2007independence}
Stephen~R. Dixon, Christopher~D. Wickens, and Jason~S. McCarley.
\newblock On the independence of compliance and reliance: Are automation false
  alarms worse than misses?
\newblock \emph{Human Factors}, 49\penalty0 (4):\penalty0 564--572, 2007.

\bibitem[Efron(1979)]{efron1979bootstrap}
Bradley Efron.
\newblock Bootstrap methods: Another look at the jackknife.
\newblock \emph{The Annals of Statistics}, 7\penalty0 (1):\penalty0 1--26,
  1979.

\bibitem[Flaborea et~al.(2024)Flaborea, D'Amely~di Melendugno, Plini, Scofano,
  De~Matteis, Furnari, Farinella, and Galasso]{flaborea2024prego}
Alessandro Flaborea, Guido~Maria D'Amely~di Melendugno, Leonardo Plini, Luca
  Scofano, Edoardo De~Matteis, Antonino Furnari, Giovanni~Maria Farinella, and
  Fabio Galasso.
\newblock {PREGO}: Online mistake detection in {PR}ocedural {EGO}centric
  videos.
\newblock In \emph{Proceedings of the IEEE/CVF Conference on Computer Vision
  and Pattern Recognition (CVPR)}, pages 18483--18492, 2024.

\bibitem[Gong et~al.(2026)Gong, Yan, Kang, Zheng, Ruan, Lu, Zhang, Sato, and
  Huang]{gong2026vinci2}
Sitong Gong, Tianyu Yan, Caixin Kang, Bo Zheng, Xiang Ruan, Huchuan Lu, Kaipeng
  Zhang, Yoichi Sato, and Yifei Huang.
\newblock {Vinci2}: Providing proactive assistance in continuous egocentric
  videos.
\newblock In \emph{Proceedings of the European Conference on Computer Vision
  (ECCV)}, pages 156--174. Springer, 2026.

\bibitem[Goyal et~al.(2017)Goyal, Khot, Summers-Stay, Batra, and
  Parikh]{goyal2017making}
Yash Goyal, Tejas Khot, Douglas Summers-Stay, Dhruv Batra, and Devi Parikh.
\newblock Making the {V} in {VQA} matter: Elevating the role of image
  understanding in visual question answering.
\newblock In \emph{Proceedings of the IEEE Conference on Computer Vision and
  Pattern Recognition (CVPR)}, pages 6325--6334, 2017.

\bibitem[Grauman et~al.(2022)Grauman, Westbury, Byrne, Chavis, Furnari,
  Girdhar, Hamburger, Jiang, Liu, Liu, Martin, Nagarajan, Radosavovic,
  Ramakrishnan, Ryan, Sharma, Wray, Xu, Xu, Zhao, Bansal, Batra, Cartillier,
  Crane, Do, Doulaty, Erapalli, Feichtenhofer, Fragomeni, Fu, Gebreselasie,
  Gonzalez, Hillis, Huang, Huang, Jia, Khoo, Kolar, Kottur, Kumar, Landini, Li,
  Li, Li, Mangalam, Modhugu, Munro, Murrell, Nishiyasu, Price, Puentes,
  Ramazanova, Sari, Somasundaram, Southerland, Sugano, Tao, Vo, Wang, Wu, Yagi,
  Zhao, Zhu, Arbelaez, Crandall, Damen, Farinella, Fuegen, Ghanem, Ithapu,
  Jawahar, Joo, Kitani, Li, Newcombe, Oliva, Park, Rehg, Sato, Shi, Shou,
  Torralba, Torresani, Yan, and Malik]{grauman2022ego4d}
Kristen Grauman, Andrew Westbury, Eugene Byrne, Zachary Chavis, Antonino
  Furnari, Rohit Girdhar, Jackson Hamburger, Hao Jiang, Miao Liu, Xingyu Liu,
  Miguel Martin, Tushar Nagarajan, Ilija Radosavovic, Santhosh~Kumar
  Ramakrishnan, Fiona Ryan, Jayant Sharma, Michael Wray, Mengmeng Xu,
  Eric~Zhongcong Xu, Chen Zhao, Siddhant Bansal, Dhruv Batra, Vincent
  Cartillier, Sean Crane, Tien Do, Morrie Doulaty, Akshay Erapalli, Christoph
  Feichtenhofer, Adriano Fragomeni, Qichen Fu, Abrham Gebreselasie, Cristina
  Gonzalez, James Hillis, Xuhua Huang, Yifei Huang, Wenqi Jia, Weslie Khoo,
  Jachym Kolar, Satwik Kottur, Anurag Kumar, Federico Landini, Chao Li, Yanghao
  Li, Zhenqiang Li, Karttikeya Mangalam, Raghava Modhugu, Jonathan Munro,
  Tullie Murrell, Takumi Nishiyasu, Will Price, Paola~Ruiz Puentes, Merey
  Ramazanova, Leda Sari, Kiran Somasundaram, Audrey Southerland, Yusuke Sugano,
  Ruijie Tao, Minh Vo, Yuchen Wang, Xindi Wu, Takuma Yagi, Ziwei Zhao, Yunyi
  Zhu, Pablo Arbelaez, David Crandall, Dima Damen, Giovanni~Maria Farinella,
  Christian Fuegen, Bernard Ghanem, Vamsi~Krishna Ithapu, C.~V. Jawahar,
  Hanbyul Joo, Kris Kitani, Haizhou Li, Richard Newcombe, Aude Oliva, Hyun~Soo
  Park, James~M. Rehg, Yoichi Sato, Jianbo Shi, Mike~Zheng Shou, Antonio
  Torralba, Lorenzo Torresani, Mingfei Yan, and Jitendra Malik.
\newblock {Ego4D}: Around the world in 3,000 hours of egocentric video.
\newblock In \emph{Proceedings of the IEEE/CVF Conference on Computer Vision
  and Pattern Recognition (CVPR)}, pages 18973--18990, 2022.

\bibitem[Guo et~al.(2026{\natexlab{a}})Guo, Chen, Zhu, Ji, Jiang, Zhong, and
  Zhang]{guo2026pace}
Siqi Guo, Yuxia Chen, Wenbo Zhu, Yangguang Ji, Zhenxiang Jiang, Teng Zhong, and
  Yu Zhang.
\newblock {PACE}: Plan, attend, check, and emit for training-free response
  timing in streaming video understanding, 2026{\natexlab{a}}.
\newblock arXiv preprint arXiv:2610.28323.

\bibitem[Guo et~al.(2026{\natexlab{b}})Guo, Pu, and Kong]{guo2026procedural}
Wenliang Guo, Yujiang Pu, and Yu Kong.
\newblock Procedural mistake detection via action effect modeling.
\newblock In \emph{International Conference on Learning Representations},
  2026{\natexlab{b}}.

\bibitem[Haneji et~al.(2025)Haneji, Nishimura, Kameko, Shirai, Yoshida,
  Kajimura, Yamamoto, Cui, Nishimoto, and Mori]{haneji2025egoops}
Yuto Haneji, Taichi Nishimura, Hirotaka Kameko, Keisuke Shirai, Tomoya Yoshida,
  Keiya Kajimura, Koki Yamamoto, Taiyu Cui, Tomohiro Nishimoto, and Shinsuke
  Mori.
\newblock {EgoOops}: A dataset for mistake action detection from egocentric
  videos referring to procedural texts.
\newblock In \emph{Proceedings of the IEEE/CVF International Conference on
  Computer Vision (ICCV) Workshops}, pages 2711--2721, 2025.

\bibitem[Holm(1979)]{holm1979simple}
Sture Holm.
\newblock A simple sequentially rejective multiple test procedure.
\newblock \emph{Scandinavian Journal of Statistics}, 6\penalty0 (2):\penalty0
  65--70, 1979.

\bibitem[Hu et~al.(2022)Hu, Shen, Wallis, Allen-Zhu, Li, Wang, Wang, and
  Chen]{hu2022lora}
Edward~J. Hu, Yelong Shen, Phillip Wallis, Zeyuan Allen-Zhu, Yuanzhi Li, Shean
  Wang, Lu Wang, and Weizhu Chen.
\newblock {LoRA}: Low-rank adaptation of large language models.
\newblock In \emph{International Conference on Learning Representations}, 2022.

\bibitem[Huang et~al.(2025)Huang, Li, Xia, Tang, Lin, Hu, and
  Zheng]{huang2025modeling}
Wei-Jin Huang, Yuan-Ming Li, Zhi-Wei Xia, Yu-Ming Tang, Kun-Yu Lin, Jian-Fang
  Hu, and Wei-Shi Zheng.
\newblock Modeling multiple normal action representations for error detection
  in procedural tasks.
\newblock In \emph{Proceedings of the IEEE/CVF Conference on Computer Vision
  and Pattern Recognition (CVPR)}, pages 27794--27804, 2025.

\bibitem[Kim(2026)]{kim2026egovoice}
Heeseung Kim.
\newblock {EgoVoice}: Proactive spoken assistance from egocentric multimodal
  streams.
\newblock In \emph{Proceedings of the 2026 Conference on Empirical Methods in
  Natural Language Processing}, pages 29112--29136, Budapest, Hungary, 2026.
  Association for Computational Linguistics.

\bibitem[Kundu et~al.(2026)Kundu, Shrivastava, Arap, Wang, Zhu, Fettes, Tiwari,
  Suresh, Singhi, Moutakanni, Castillejo~Munoz, Bolourchi, Fung, Donmez,
  Damavandi, Kumar, and Moon]{kundu2026plan}
Kaustav Kundu, Ritvik Shrivastava, Maxim Arap, Nanshu Wang, Xianhui Zhu,
  Quintin Fettes, Gautam Tiwari, Parth Suresh, Prateek Singhi, Th{\'e}o
  Moutakanni, Alejandro Castillejo~Munoz, Allen Bolourchi, Pascale Fung, Pinar
  Donmez, Babak Damavandi, Anuj Kumar, and Seungwhan Moon.
\newblock Plan, watch, recover: A benchmark and architectures for proactive
  procedural assistance, 2026.
\newblock arXiv preprint arXiv:2606.04970.

\bibitem[Lee et~al.(2024)Lee, Lu, Zhang, Hoai, and Elhamifar]{lee2024error}
Shih-Po Lee, Zijia Lu, Zekun Zhang, Minh Hoai, and Ehsan Elhamifar.
\newblock Error detection in egocentric procedural task videos.
\newblock In \emph{Proceedings of the IEEE/CVF Conference on Computer Vision
  and Pattern Recognition (CVPR)}, pages 18655--18666, 2024.

\bibitem[Li et~al.(2025{\natexlab{a}})Li, Zhang, Guo, Zhang, Li, Zhang, Zhang,
  Zhang, Li, Liu, and Li]{li2025llavaonevision}
Bo Li, Yuanhan Zhang, Dong Guo, Renrui Zhang, Feng Li, Hao Zhang, Kaichen
  Zhang, Peiyuan Zhang, Yanwei Li, Ziwei Liu, and Chunyuan Li.
\newblock {LLaVA-OneVision}: Easy visual task transfer.
\newblock \emph{Transactions on Machine Learning Research}, 2025{\natexlab{a}}.

\bibitem[Li et~al.(2025{\natexlab{b}})Li, Xu, Cheng, Wu, Yap, Chau, and
  Wang]{li2025building}
Junlong Li, Huaiyuan Xu, Sijie Cheng, Kejun Wu, Kim-Hui Yap, Lap-Pui Chau, and
  Yi Wang.
\newblock Building egocentric procedural {AI} assistant: Methods, benchmarks,
  and challenges, 2025{\natexlab{b}}.
\newblock arXiv preprint arXiv:2511.13261.

\bibitem[Li et~al.(2025{\natexlab{c}})Li, Hu, Shao, Shen, and
  Nie]{li2025lionfs}
Wei Li, Bing Hu, Rui Shao, Leyang Shen, and Liqiang Nie.
\newblock {LION-FS}: Fast \& slow video-language thinker as online video
  assistant.
\newblock In \emph{Proceedings of the IEEE/CVF Conference on Computer Vision
  and Pattern Recognition (CVPR)}, pages 3240--3251, 2025{\natexlab{c}}.

\bibitem[Li et~al.(2026)Li, Jain, Bellos, and Corso]{li2026mistake}
Yayuan Li, Aadit Jain, Filippos Bellos, and Jason~J. Corso.
\newblock Mistake attribution: Fine-grained mistake understanding in egocentric
  videos.
\newblock In \emph{Proceedings of the IEEE/CVF Conference on Computer Vision
  and Pattern Recognition (CVPR)}, pages 23966--23976, 2026.

\bibitem[Liu et~al.(2026)Liu, Chen, Xu, Liu, Guan, Lu, Bo, Hancke, Liu, and
  Lau]{liu2026guideme}
Fang Liu, Jinpeng Chen, Ke Xu, Yuhao Liu, Huankang Guan, Xudong Lu, Yang Bo,
  Gerhard Hancke, Rui Liu, and Rynson W.~H. Lau.
\newblock {GuideMe}: Multi-domain task guidance and intervention in streaming
  video.
\newblock In \emph{Proceedings of the European Conference on Computer Vision
  (ECCV)}, pages 57--74. Springer, 2026.

\bibitem[Ozsoy et~al.(2026)Ozsoy, Doorenbos, Spurio, Francesca, and
  Gall]{ozsoy2026unreasonable}
Serdar Ozsoy, Lars Doorenbos, Federico Spurio, Gianpiero Francesca, and Juergen
  Gall.
\newblock The unreasonable effectiveness of {VLMs} for zero-shot procedural
  mistake detection, 2026.
\newblock arXiv preprint arXiv:2606.21579.

\bibitem[Peddi et~al.(2024)Peddi, Arya, Challa, Pallapothula, Vyas, Gouripeddi,
  Zhang, Wang, Komaragiri, Ragan, Ruozzi, Xiang, and
  Gogate]{peddi2024captaincook4d}
Rohith Peddi, Shivvrat Arya, Bharath Challa, Likhitha Pallapothula, Akshay
  Vyas, Bhavya Gouripeddi, Qifan Zhang, Jikai Wang, Vasundhara Komaragiri, Eric
  Ragan, Nicholas Ruozzi, Yu Xiang, and Vibhav Gogate.
\newblock {CaptainCook4D}: A dataset for understanding errors in procedural
  activities.
\newblock In \emph{Advances in Neural Information Processing Systems (Datasets
  and Benchmarks Track)}, pages 135626--135679, 2024.

\bibitem[Pu et~al.(2025)Pu, Zhang, Sendhilnathan, Freitag, Sodhi, and
  Jonker]{pu2025promemassist}
Kevin Pu, Ting Zhang, Naveen Sendhilnathan, Sebastian Freitag, Raj Sodhi, and
  Tanya~R. Jonker.
\newblock {ProMemAssist}: Exploring timely proactive assistance through working
  memory modeling in multi-modal wearable devices.
\newblock In \emph{Proceedings of the 38th Annual ACM Symposium on User
  Interface Software and Technology (UIST)}, pages 1--19. ACM, 2025.

\bibitem[Qian et~al.(2025)Qian, Ding, Dong, Zhang, Zang, Cao, Lin, and
  Wang]{qian2025dispider}
Rui Qian, Shuangrui Ding, Xiaoyi Dong, Pan Zhang, Yuhang Zang, Yuhang Cao,
  Dahua Lin, and Jiaqi Wang.
\newblock Dispider: Enabling video {LLMs} with active real-time interaction via
  disentangled perception, decision, and reaction.
\newblock In \emph{Proceedings of the IEEE/CVF Conference on Computer Vision
  and Pattern Recognition (CVPR)}, pages 24045--24055, 2025.

\bibitem[Seminara et~al.(2024)Seminara, Farinella, and
  Furnari]{seminara2024differentiable}
Luigi Seminara, Giovanni~Maria Farinella, and Antonino Furnari.
\newblock Differentiable task graph learning: Procedural activity
  representation and online mistake detection from egocentric videos.
\newblock In \emph{Advances in Neural Information Processing Systems}, pages
  59373--59407, 2024.

\bibitem[Sener et~al.(2022)Sener, Chatterjee, Shelepov, He, Singhania, Wang,
  and Yao]{sener2022assembly101}
Fadime Sener, Dibyadip Chatterjee, Daniel Shelepov, Kun He, Dipika Singhania,
  Robert Wang, and Angela Yao.
\newblock {Assembly101}: A large-scale multi-view video dataset for
  understanding procedural activities.
\newblock In \emph{Proceedings of the IEEE/CVF Conference on Computer Vision
  and Pattern Recognition (CVPR)}, pages 21096--21106, 2022.

\bibitem[Spurio et~al.(2026)Spurio, Zatsarynna, Doorenbos, Bahrami, Francesca,
  and Gall]{spurio2026posttraining}
Federico Spurio, Olga Zatsarynna, Lars Doorenbos, Emad Bahrami, Gianpiero
  Francesca, and Juergen Gall.
\newblock Post-training {VLMs} for video mistake detection.
\newblock In \emph{Proceedings of the British Machine Vision Conference
  (BMVC)}, 2026.

\bibitem[Thrush et~al.(2022)Thrush, Jiang, Bartolo, Singh, Williams, Kiela, and
  Ross]{thrush2022winoground}
Tristan Thrush, Ryan Jiang, Max Bartolo, Amanpreet Singh, Adina Williams, Douwe
  Kiela, and Candace Ross.
\newblock Winoground: Probing vision and language models for visio-linguistic
  compositionality.
\newblock In \emph{Proceedings of the IEEE/CVF Conference on Computer Vision
  and Pattern Recognition (CVPR)}, pages 5238--5248, 2022.

\bibitem[Tong et~al.(2024)Tong, Liu, Zhai, Ma, LeCun, and Xie]{tong2024eyes}
Shengbang Tong, Zhuang Liu, Yuexiang Zhai, Yi Ma, Yann LeCun, and Saining Xie.
\newblock Eyes wide shut? exploring the visual shortcomings of multimodal
  {LLMs}.
\newblock In \emph{Proceedings of the IEEE/CVF Conference on Computer Vision
  and Pattern Recognition (CVPR)}, pages 9568--9578, 2024.

\bibitem[Umapathi(2026)]{umapathi2026ambient}
Logesh~Kumar Umapathi.
\newblock Ambient @ {EgoProactive} 2026: Proactive egocentric assistance with
  visually grounded supervision, 2026.
\newblock arXiv preprint arXiv:2609.07099. Winning solution technical report,
  EgoProactive track, ECCV 2026 Wearable AI Grand Challenge.

\bibitem[{VAR Workshop}(2026)]{var2026aicoach}
{VAR Workshop}.
\newblock {AI} coach challenge @ {VAR} 2026: Don't just watch. intervene.
\newblock Challenge of the workshop on Vision-based Assistants in the Real
  World, Second Edition, at CVPR 2026, 2026.

\bibitem[Wang et~al.(2025{\natexlab{a}})Wang, Feng, Lai, Xu, Li, Ge, Dehghan,
  Cao, and Huang]{wang2025streambridge}
Haibo Wang, Bo Feng, Zhengfeng Lai, Mingze Xu, Shiyu Li, Weifeng Ge, Afshin
  Dehghan, Meng Cao, and Ping Huang.
\newblock {StreamBridge}: Turning your offline video large language model into
  a proactive streaming assistant.
\newblock In \emph{Advances in Neural Information Processing Systems}, pages
  132332--132359, 2025{\natexlab{a}}.

\bibitem[Wang et~al.(2025{\natexlab{b}})Wang, Gao, Gu, Pu, Cui, Wei, Liu, Jing,
  Ye, Shao, Wang, Chen, Zhang, Yang, Wang, Wei, Yin, Li, Cui, Chen, Ding, Tian,
  Wu, Xie, Li, Yang, Duan, Wang, Hou, Hao, Zhang, Li, Zhao, Duan, Deng, Fu, He,
  Wang, He, Shi, He, Xiong, Lv, Wu, Shao, Zhang, Deng, Qi, Ge, Guo, Zhang,
  Zhang, Cao, Lin, Tang, Gao, Huang, Gu, Lyu, Tang, Wang, Lv, Ouyang, Wang,
  Dou, Zhu, Lu, Lin, Dai, Su, Zhou, Chen, Qiao, Wang, and
  Luo]{wang2025internvl35}
Weiyun Wang, Zhangwei Gao, Lixin Gu, Hengjun Pu, Long Cui, Xingguang Wei,
  Zhaoyang Liu, Linglin Jing, Shenglong Ye, Jie Shao, Zhaokai Wang, Zhe Chen,
  Hongjie Zhang, Ganlin Yang, Haomin Wang, Qi Wei, Jinhui Yin, Wenhao Li, Erfei
  Cui, Guanzhou Chen, Zichen Ding, Changyao Tian, Zhenyu Wu, Jingjing Xie,
  Zehao Li, Bowen Yang, Yuchen Duan, Xuehui Wang, Zhi Hou, Haoran Hao, Tianyi
  Zhang, Songze Li, Xiangyu Zhao, Haodong Duan, Nianchen Deng, Bin Fu, Yinan
  He, Yi Wang, Conghui He, Botian Shi, Junjun He, Yingtong Xiong, Han Lv, Lijun
  Wu, Wenqi Shao, Kaipeng Zhang, Huipeng Deng, Biqing Qi, Jiaye Ge, Qipeng Guo,
  Wenwei Zhang, Songyang Zhang, Maosong Cao, Junyao Lin, Kexian Tang, Jianfei
  Gao, Haian Huang, Yuzhe Gu, Chengqi Lyu, Huanze Tang, Rui Wang, Haijun Lv,
  Wanli Ouyang, Limin Wang, Min Dou, Xizhou Zhu, Tong Lu, Dahua Lin, Jifeng
  Dai, Weijie Su, Bowen Zhou, Kai Chen, Yu Qiao, Wenhai Wang, and Gen Luo.
\newblock {InternVL3.5}: Advancing open-source multimodal models in
  versatility, reasoning, and efficiency, 2025{\natexlab{b}}.
\newblock arXiv preprint arXiv:2508.18265.

\bibitem[Wang et~al.(2023)Wang, Kwon, Rad, Pan, Chakraborty, Andrist, Bohus,
  Feniello, Tekin, Frujeri, Joshi, and Pollefeys]{wang2023holoassist}
Xin Wang, Taein Kwon, Mahdi Rad, Bowen Pan, Ishani Chakraborty, Sean Andrist,
  Dan Bohus, Ashley Feniello, Bugra Tekin, Felipe~Vieira Frujeri, Neel Joshi,
  and Marc Pollefeys.
\newblock {HoloAssist}: an egocentric human interaction dataset for interactive
  {AI} assistants in the real world.
\newblock In \emph{Proceedings of the IEEE/CVF International Conference on
  Computer Vision (ICCV)}, pages 20270--20281, 2023.

\bibitem[Wang et~al.(2025{\natexlab{c}})Wang, Meng, Wang, Liang, Wei, Zhang,
  and Zhao]{wang2025videollm}
Yueqian Wang, Xiaojun Meng, Yuxuan Wang, Jianxin Liang, Jiansheng Wei, Huishuai
  Zhang, and Dongyan Zhao.
\newblock {VideoLLM} knows when to speak: Enhancing time-sensitive video
  comprehension with video-text duet interaction format.
\newblock In \emph{Findings of the Association for Computational Linguistics:
  {EMNLP} 2025}, pages 6338--6359, Suzhou, China, 2025{\natexlab{c}}.
  Association for Computational Linguistics.

\bibitem[Wang et~al.(2025{\natexlab{d}})Wang, Meng, Wang, Zhang, and
  Zhao]{wang2025proactivevideoqa}
Yueqian Wang, Xiaojun Meng, Yifan Wang, Huishuai Zhang, and Dongyan Zhao.
\newblock {ProactiveVideoQA}: A comprehensive benchmark evaluating proactive
  interactions in video large language models, 2025{\natexlab{d}}.
\newblock arXiv preprint arXiv:2507.09313.

\bibitem[Wang et~al.(2026)Wang, Liu, Wang, Xu, Wan, Zhang, and
  Zhao]{wang2026mmduet2}
Yueqian Wang, Songxiang Liu, Disong Wang, Nuo Xu, Guanglu Wan, Huishuai Zhang,
  and Dongyan Zhao.
\newblock {MMDuet2}: Enhancing proactive interaction of video {MLLMs} with
  multi-turn reinforcement learning.
\newblock In \emph{International Conference on Learning Representations}, 2026.

\bibitem[Wen et~al.(2026)Wen, Yang, Weissert, Scherrer, Z{\"o}llner, Liu, Chen,
  Wei, Zheng, and Peng]{wen2026provia}
Di Wen, Kailun Yang, Jimmy Weissert, Luc~Maria Scherrer, Cedric Z{\"o}llner,
  Ruiping Liu, Yufan Chen, Jiale Wei, Junwei Zheng, and Kunyu Peng.
\newblock {PROVIA}: Procedure state tracking for online mistake detection in
  egocentric videos, 2026.
\newblock arXiv preprint arXiv:2609.20638.

\bibitem[Xu et~al.(2026)Xu, Yang, Jiang, Liu, Hou, Fan, Chen, Yan, and
  Jiang]{xu2026pro2assist}
Lilin Xu, Bufang Yang, Siyang Jiang, Kaiwei Liu, Kaiyuan Hou, Yuang Fan,
  Hongkai Chen, Zhenyu Yan, and Xiaofan Jiang.
\newblock {Pro$^{2}$Assist}: Continuous step-aware proactive assistance with
  multi-modal egocentric perception for long-horizon procedural tasks.
\newblock \emph{Proceedings of the ACM on Interactive, Mobile, Wearable and
  Ubiquitous Technologies}, 10\penalty0 (3):\penalty0 1--31, 2026.

\bibitem[Yan et~al.(2026)Yan, Dai, Ran, Li, Lin, Jin, Xie, Liao, and
  Lian]{yan2026proactvl}
Weicai Yan, Yuhong Dai, Qi Ran, Haodong Li, Wang Lin, Tao Jin, Xing Xie, Hao
  Liao, and Jianxun Lian.
\newblock {Proact-VL}: A proactive {VideoLLM} for real-time {AI} companions.
\newblock In \emph{Proceedings of the 43rd International Conference on Machine
  Learning}, pages 143579--143618. PMLR, 2026.

\bibitem[Yang et~al.(2026)Yang, Tang, Zhao, Zhuang, Lu, An, Hu, Zhang, Swikir,
  He, Ge, Khan, and Razzak]{yang2026anticipatory}
Haolin Yang, Feilong Tang, Lingxiao Zhao, Xinlin Zhuang, Yifan Lu, Xiang An,
  Ming Hu, Xiaofeng Zhang, Abdalla Swikir, Junjun He, Zongyuan Ge,
  Muhammad~Haris Khan, and Imran Razzak.
\newblock Towards anticipatory agents for streaming video understanding.
\newblock In \emph{Findings of the Association for Computational Linguistics:
  {EMNLP} 2026}, pages 1315--1332, Budapest, Hungary, 2026. Association for
  Computational Linguistics.

\bibitem[Yang et~al.(2025)Yang, Zhang, Hu, Wang, Qian, Wen, Yang, Gao, Dong,
  and Xu]{yang2025livestar}
Zhenyu Yang, Kairui Zhang, Yuhang Hu, Bing Wang, Shengsheng Qian, Bin Wen, Fan
  Yang, Tingting Gao, Weiming Dong, and Changsheng Xu.
\newblock {LiveStar}: Live streaming assistant for real-world online video
  understanding.
\newblock In \emph{Advances in Neural Information Processing Systems}, pages
  31266--31304, 2025.

\bibitem[Zhang et~al.(2025{\natexlab{a}})Zhang, Dong, Lin, Madotto, Kumar,
  Damavandi, Chai, and Moon]{zhang2025proactive}
Yichi Zhang, Xin~Luna Dong, Zhaojiang Lin, Andrea Madotto, Anuj Kumar, Babak
  Damavandi, Joyce Chai, and Seungwhan Moon.
\newblock Proactive assistant dialogue generation from streaming egocentric
  videos.
\newblock In \emph{Proceedings of the 2025 Conference on Empirical Methods in
  Natural Language Processing}, pages 12044--12068, Suzhou, China,
  2025{\natexlab{a}}. Association for Computational Linguistics.

\bibitem[Zhang et~al.(2025{\natexlab{b}})Zhang, Shi, Wang, and
  Yang]{zhang2025eyes}
Yulin Zhang, Cheng Shi, Yang Wang, and Sibei Yang.
\newblock Eyes wide open: Ego proactive video-{LLM} for streaming video.
\newblock In \emph{Advances in Neural Information Processing Systems}, pages
  13420--13463, 2025{\natexlab{b}}.

\bibitem[Zheng et~al.(2026)Zheng, Ding, Yang, Jiang, Wu, Zhang, Wang, Cao, and
  Liu]{zheng2026emgarde}
Yikai Zheng, Xin Ding, Yifan Yang, Shiqi Jiang, Hao Wu, Qianxi Zhang, Weijun
  Wang, Ting Cao, and Yunxin Liu.
\newblock {Em-Garde}: A propose-match framework for proactive streaming video
  understanding.
\newblock In \emph{Proceedings of the European Conference on Computer Vision
  (ECCV)}, pages 504--521. Springer, 2026.

\end{thebibliography}
\clearpage
\endgroup
\appendix
\setcounter{figure}{0}
\setcounter{table}{0}
\renewcommand{\thefigure}{S\arabic{figure}}
\renewcommand{\thetable}{S\arabic{table}}
\renewcommand{\theHfigure}{supp.\arabic{figure}}
\renewcommand{\theHtable}{supp.\arabic{table}}
\twocolumn[{\begin{center}\pdfbookmark[0]{Supplementary Material}{preprint.supplement}\Large\bfseries Supplementary Material\end{center}\vspace{0.5em}}]
\noindent Paired negatives change \emph{when} ECT responds: under the official protocol, it confirms steps on time more often and raises fewer false alarms, at the cost of some mistake recall.

\smallskip\noindent\textbf{Reading guide.} \Cref{sup:protocol,sup:frontier} ask how much stalling and video-blind timing can explain. \Cref{sup:results,sup:operating} compare the training variants and decoding choices. \Cref{sup:pairs,sup:decomp,sup:content,sup:perturb} examine individual decisions, feedback and video perturbations. \Cref{sup:code} describes the scope of the public materials. The main paper and supplement share one list of \hyperref[sec:references]{references}. Analyses marked $^\dagger$ were designed after the test results; they change no reported model or main result, and \cref{sup:operating} lists the use of the test sets in order.

\section{Protocol, models and settings}
\label{sup:protocol}

\paragraph{Protocol.}
We use the official per-step evaluation loop and scorer for both benchmarks. Inference settings and any changes to prompting or answer parsing are described below.
The model is given the ground-truth instruction of the current step; a check is made every 10 frames (5\,s at 2 frames per second) on the frames since the annotated start, at most 240 (120\,s).
In the protocol described in \cref{sec:diagnosis}, the annotated start anchors the window, but only a model confirmation advances the instruction. If a step is never confirmed, the model keeps being checked on later footage under that instruction and subsequent steps are never reached. In four audited runs on the 36 QICD test recordings, the number of instructions issued equaled the number of confirmations plus one, capped at the number of steps.
At each check the completion question is asked first and, if the answer is negative, the mistake question.
An answer is positive if it contains ``yes''; our fine-tuned models answer ``YES'' or ``NO'' at the start of the output, so the rule is unambiguous for them.
A mistake-free step contributes an IC hit only if its single prediction is a confirmation within the 15\,s tolerance of the annotated step end; any mistake feedback during the step voids the hit. Unreached mistake-free steps remain in the denominator.
The memory prompt is the one-minute feedback memory of \citet{bhattacharyya2026streaming}; in our runs the memory is cleared at every step.
Of the 40 sessions of Ego-MC-Bench, 39 are released (562 instructions, 397 mistake annotations); the published results use 40.

\paragraph{Stalling and how failed steps are annotated$^\dagger$.}
The two benchmarks annotate a step with a mistake differently. In QICD Advanced Planning, 236 of the 474 test steps contain a mistake and none of them carries an annotated success: the next step begins after the mistake, in 65 cases with the same instruction. In Ego-MC-Bench, 282 of the 298 steps with a mistake end with a success after the correction, and 16 (3\% of all steps) end with the mistake.
Since only a confirmation advances the instruction, a model that withholds confirmation on a failed QICD step stays on it.
Among the recordings that a run leaves before the last step, the step where it stops ends with a mistake in 7 of 14 for zero-shot Qwen3-VL-2B on QICD, 11 of 24 for Qwen3.5-397B and 27 of 36 for Qwen2.5-VL-7B; on Ego-MC-Bench the same models stop at such a step in 0 of 5, 0 of 16 and 1 of 38 recordings, and otherwise at a mistake-free step or after a corrected mistake. The greedy ECT seeds stop early in 0--4 recordings per test set.
Recomputing the IC denominator over the mistake-free steps each run reaches, with the official numerator, zero-shot Qwen3-VL-2B with the memory prompt reaches 385/474 steps on QICD and its IC rises from 10.5 to 13.5; Qwen3.5-397B rises from 18.1 to 28.7 on QICD and from 34.5 to 46.9 on Ego-MC-Bench; no greedy ECT seed changes by more than two points. Zero-shot Qwen2.5-VL-7B reaches only 96/562 steps on Ego-MC-Bench, and its IC on that subset is 58.3 against 8.0 officially. Stalling thus explains part of the official gap of the zero-shot models, and not the low IC of the main zero-shot reference. Each reached-only IC conditions on a model-specific subset; the counts are released with the code.

\paragraph{Zero-shot models.}
\Cref{tab:zeroshot} lists every zero-shot run.
InternVL3.5~\cite{wang2025internvl35} and LLaVA-OneVision~\cite{li2025llavaonevision} receive 32 frames sampled uniformly from the window, zero-shot and fine-tuned alike.
The Qwen3.5 and Qwen3.8 models are run through a hosted inference service with 16 frames per check (448 pixels wide, with time stamps), the memory prompt, temperature 0 and reasoning disabled.
For these models we append ``Start your answer with YES or NO, then give at most one sentence.'' to the question and parse only the leading word, because without the suffix some answers contain ``yes'' mid-sentence.
Requests rejected by the service's content filter are counted as ``NO''; this affects at most 108 checks per model on QICD and none on Ego-MC-Bench.
Two models were accidentally queried twice: the decisions of the two passes agree in 99.9--100\% of the checks.
With these settings Qwen3.5-27B reaches 42.0 / 0.068 on Ego-MC-Bench, below the 45.5 / 0.14 published for it with its own settings~\cite{bhattacharyya2026streaming}; the paper uses the published value when it discusses published results.
The zero-shot conclusion does not depend on greedy decoding: with sampling, Qwen3-VL-2B reaches 5.5 / 0.099, 3.8 / 0.103 and 2.9 / 0.102 on QICD over three seeds.

\begin{table*}[t]
  \centering
  \small
  \setlength{\tabcolsep}{4pt}
  \resizebox{\ifdim\width>\linewidth\linewidth\else\width\fi}{!}{%
  \begin{tabular}{@{}llcccc@{}}
    \toprule
    Model & Setting & QICD & $\Delta$ & Ego-MC-Bench & $\Delta$ \\
    \midrule
    Qwen3-VL-2B & official prompt, up to 240 frames & 4.2 / 0.074 & $-0.243$ & 14.0 / 0.147 & $-0.195$ \\
    Qwen3-VL-2B & + memory prompt & 10.5 / 0.146 & $-0.170$ & 14.4 / 0.215 & $-0.124$ \\
    Qwen3-VL-8B & + memory prompt & 2.9 / 0.000 & $-0.317$ & -- & -- \\
    Qwen2.5-VL-7B & + memory prompt & 1.7 / 0.000 & $-0.317$ & 8.0 / 0.000 & $-0.365$ \\
    InternVL3.5-2B & 32 frames & 32.8 / 0.000 & $-0.230$ & 49.2 / 0.000 & $-0.114$ \\
    InternVL3.5-8B & 32 frames & 3.4 / 0.000 & $-0.317$ & 11.4 / 0.000 & $-0.353$ \\
    LLaVA-OneVision-7B & 32 frames & 0.0 / 0.000 & $-0.317$ & 0.0 / 0.000 & $-0.373$ \\
    \midrule
    Qwen3.5-27B & 16 frames, see text & 16.8 / 0.026 & $-0.283$ & 42.0 / 0.068 & $-0.105$ \\
    Qwen3.5-35B-A3B & 16 frames, see text & 3.4 / 0.000 & $-0.317$ & 6.4 / 0.000 & $-0.371$ \\
    Qwen3.5-122B-A10B & 16 frames, see text & 13.0 / 0.069 & $-0.245$ & 29.5 / 0.066 & $-0.182$ \\
    Qwen3.5-397B-A17B & 16 frames, see text & 18.1 / 0.203 & $-0.104$ & 34.5 / 0.239 & $+0.021$ \\
    Qwen3.8-Omni-Flash & 16 frames, see text & 8.0 / 0.075 & $-0.241$ & 29.5 / 0.112 & $-0.136$ \\
    Qwen3.8-Max & 16 frames, see text & 23.9 / 0.189 & $-0.090$ & 37.9 / 0.324 & $+0.126$ \\
    \bottomrule
  \end{tabular}}
  \caption{All zero-shot runs under the official per-step protocol with greedy decoding (IC-Acc / mistake F1 and margin over the frontier). The twelve zero-shot models of the paper are the twelve distinct models of this table.}
  \label{tab:zeroshot}
\end{table*}

\section{The video-blind frontier}
\label{sup:frontier}

\paragraph{Construction.}
The rules use the step start, the elapsed time, the instruction text and statistics of the QICD training split.
(i)~Fixed delays: confirm $A$ seconds into the step; optionally raise one alarm at $B$ seconds, either in every step or in a fixed fraction of the steps.
(ii)~Duration priors: confirm at a multiple of the median training duration of the same instruction (the global median if the instruction is unseen).
(iii)~Mistake priors: raise the alarm only for instructions whose mistake rate in the training split exceeds a threshold.
Every rule and every per-step mixture along the hull is scored with the official evaluation code; \cref{tab:vertices} gives the vertices.
The main text adds to this hull the learned blind policy on the same inputs described below (``paper's inputs''); its margins are against that combined frontier. Margins in this supplement are against the rules alone unless stated, and \cref{tab:blindstrong} compares the readings.
If the two delays of the fixed-delay rule are instead chosen on the QICD validation split (alarm at 15\,s, confirm at 20\,s), the rule reaches a mistake F1 of 0.361 on Ego-MC-Bench.
Above the highest completion accuracy of any blind rule the frontier is zero, so the margin equals the mistake F1 there; the paper marks these cases.

\begin{table*}[t]
  \centering
  \small
  \setlength{\tabcolsep}{4pt}
  \resizebox{\ifdim\width>\linewidth\linewidth\else\width\fi}{!}{%
  \begin{tabular}{@{}ll@{}}
    \toprule
    Benchmark & Vertices of the frontier (IC-Acc, mistake F1) \\
    \midrule
    QICD & (0.0, 0.316), (4.2, 0.317), (11.8, 0.316), (18.5, 0.306), (21.0, 0.294), (23.9, 0.279), (38.7, 0.198), (44.1, 0.105), (46.6, 0.000) \\
    Ego-MC-Bench & (0.0, 0.373), (5.7, 0.373), (13.3, 0.346), (44.7, 0.157), (48.5, 0.126), (56.1, 0.000) \\
    \bottomrule
  \end{tabular}}
  \caption{Vertices of the video-blind frontier used for every margin in the paper (upper convex hull of measured blind policies, tuned on the test set).}
  \label{tab:vertices}
\end{table*}

\paragraph{A learned blind policy and resampling.}
The frontier is a lower bound on what blind policies can do.
As one check, we train a logistic regression on the 52,281 checks of the training split from elapsed time, step index, instruction text and training priors, and threshold its two outputs.
This logistic regression is weaker than the hand-written rules: its best mistake F1 is 0.131 on QICD and 0.175 on Ego-MC-Bench, and its best completion accuracy 38.7 and 46.6; a stronger learner is not, as the next paragraph shows.
\Cref{tab:boot} resamples the test videos and recomputes a frontier each time.
This frontier is the hull of 625 single policies (the rules and the learned policy) without the per-step mixtures of the frontier above, so its margins differ slightly from those of the paper (for ECT seed 0, $+0.112$ against $+0.062$).

\begin{table*}[t]
  \centering
  \small
  \setlength{\tabcolsep}{4pt}
  \resizebox{\ifdim\width>\linewidth\linewidth\else\width\fi}{!}{%
  \begin{tabular}{@{}lcc@{}}
    \toprule
    Run & QICD & Ego-MC-Bench \\
    \midrule
    ECT, seed 0 & 0.914 \; [-0.025, +0.224] & 1.000 \; [+0.153, +0.268] \\
    ECT, seed 1 & 0.996 \; [+0.021, +0.269] & 1.000 \; [+0.136, +0.257] \\
    ECT, seed 2 & 0.965 \; [-0.005, +0.241] & 1.000 \; [+0.214, +0.329] \\
    ECT, seed 3 & 0.959 \; [-0.010, +0.232] & 1.000 \; [+0.200, +0.308] \\
    ECT, seed 0, thresholds & 1.000 \; [+0.079, +0.308] & 1.000 \; [+0.137, +0.240] \\
    ECT, Qwen3-VL-4B & 0.896 \; [-0.033, +0.217] & 1.000 \; [+0.121, +0.232] \\
    + clean, seed 0 & 1.000 \; [+0.079, +0.322] & 1.000 \; [+0.146, +0.261] \\
    + notdone, seed 0 & 0.777 \; [-0.048, +0.109] & 1.000 \; [+0.186, +0.295] \\
    No neg., re-balanced sampling & 0.958 \; [-0.006, +0.130] & 1.000 \; [+0.185, +0.301] \\
    No paired neg., seed 0 & 0.183 \; [-0.087, +0.028] & 1.000 \; [+0.213, +0.320] \\
    Qwen3-VL-2B, zero-shot & 0.000 \; [-0.230, -0.123] & 0.000 \; [-0.189, -0.040] \\
    \bottomrule
  \end{tabular}}
  \caption{Probability of lying beyond the frontier when the test videos are resampled (2,000 draws; the frontier is recomputed on every draw from 625 blind policies, 49 of them learned), with the 95\% interval of the margin. --: not computed.}
  \label{tab:boot}
\end{table*}

\paragraph{A stronger learned blind policy$^\dagger$.}
Gradient-boosted trees predict each step's duration and mistake probability. They are fit on the QICD training and validation splits, which the official loader combines, without test recordings. A policy confirms at a multiple of the predicted duration plus an offset and raises one alarm at another such time if the predicted mistake probability exceeds a threshold, giving 18,900 policies.
With the inputs of the paper's definition (instruction text, step index, elapsed time and training priors), and more so when it also sees the position in the plan and the observed durations of the earlier steps of the recording, which the protocol reveals, this learner extends the frontier: tuned on the test set like the rules, its best mistake F1 rises from 0.317 to 0.353 and 0.361 on QICD and from 0.219 to 0.232 and 0.258 on the Main set.
Every policy on the new hull was re-scored with the official evaluation code, and the hand-written rules, re-implemented in the same scorer, reproduce the hull of the single rules on all three test sets to within 0.001 in F1.
The extension persists when policies are selected by their scores on one half of the test recordings and evaluated on the other. The candidate alarm thresholds are quantiles of predictions on the full test set, computed without its labels.
\Cref{tab:blindstrong} gives the margins of ECT under all three readings.
On QICD Advanced Planning, the margins of ECT with greedy decoding shrink from $+0.062$--$+0.107$ to $+0.012$--$+0.061$ with the paper's inputs (the frontier of the main text) and to $-0.028$--$+0.029$ with all inputs. Held out, a seed is beyond in 87--97\% of the 500 splits with the paper's inputs (99--100\% on Ego-MC-Bench) and in 70--96\% with all inputs, and the mean margin of every seed stays positive.
The thresholded variant remains beyond under every reading ($+0.032$ or more when tuned on the test set).
On the Main set the pattern is reversed: greedy decoding remains beyond ($+0.025$--$+0.125$ with all inputs, held out with probability 0.97--1.00), while the margins of the thresholded variant, which were $+0.022$--$+0.050$, fall to about zero with all inputs.
On Ego-MC-Bench the margins of greedy decoding hardly change ($+0.162$--$+0.278$); one thresholded seed falls just inside the frontier ($-0.026$ with all inputs).
Of the 104 results with a margin in our tables (including the published results on the Main set), none of the 40 inside the frontier of the paper moves out of it; of the 64 beyond it, 58 stay beyond with the paper's inputs and 50 with all inputs.
With all inputs, Qwen3.5-397B-A17B on Ego-MC-Bench moves inside ($+0.021$ to $-0.084$), Qwen3.8-Max stays just beyond ($+0.126$ to $+0.014$), and on the Main set LiveMamba stays just beyond ($+0.056$ to $+0.014$) and the challenge winner well beyond ($+0.250$).
Every ECT seed is beyond the hand-written frontier on both benchmarks. Against the all-input learned frontier, each test set has at least one ECT decoding that remains beyond the frontier, but neither decoding does so on all three test sets. The QICD margins are small.

\begin{table*}[t]
  \centering\small\setlength{\tabcolsep}{4pt}
  \begin{tabular*}{\linewidth}{@{\extracolsep{\fill}}llcccccc@{}}
    \toprule
    & & & \multicolumn{3}{c}{Tuned on the whole test set} & \multicolumn{2}{c}{Held out (500 half splits)} \\
    \cmidrule(lr){4-6}\cmidrule(l){7-8}
    Test set & Decoding & Seed & Rules & \shortstack{+ learned, paper's\\inputs (main text)} & \shortstack{+ learned\\all inputs} & \shortstack{Margin\\all inputs} & $P(\mathrm{margin}>0)$ \\
    \midrule
    QICD Adv.\ Planning & Greedy & 0 & $+0.062$ & $+0.012$ & \boldmath$-0.028$ & $+0.035$ & 0.704 \\
     &  & 1 & $+0.107$ & $+0.061$ & $+0.029$ & $+0.086$ & 0.956 \\
     &  & 2 & $+0.083$ & $+0.038$ & $+0.009$ & $+0.066$ & 0.878 \\
     &  & 3 & $+0.087$ & $+0.034$ & \boldmath$-0.011$ & $+0.050$ & 0.794 \\
    \addlinespace[2pt]
     & Thresholded & 0 & $+0.165$ & $+0.113$ & $+0.069$ & $+0.131$ & 0.996 \\
     &  & 1 & $+0.167$ & $+0.117$ & $+0.077$ & $+0.135$ & 0.998 \\
     &  & 2 & $+0.155$ & $+0.104$ & $+0.061$ & $+0.122$ & 0.998 \\
     &  & 3 & $+0.137$ & $+0.079$ & $+0.032$ & $+0.089$ & 0.944 \\
    \midrule
    QICD Main & Greedy & 0 & $+0.113$ & $+0.104$ & $+0.067$ & $+0.126$ & 0.998 \\
     &  & 1 & $+0.072$ & $+0.072$ & $+0.025$ & $+0.054$ & 0.966 \\
     &  & 2 & $+0.077$ & $+0.077$ & $+0.035$ & $+0.081$ & 0.990 \\
     &  & 3 & $+0.181$ & $+0.143$ & $+0.125$ & $+0.164$ & 0.998 \\
    \addlinespace[2pt]
     & Thresholded & 0 & $+0.022$ & $+0.022$ & \boldmath$-0.027$ & \boldmath$-0.001$ & 0.472 \\
     &  & 1 & $+0.038$ & $+0.038$ & \boldmath$-0.010$ & $+0.014$ & 0.704 \\
     &  & 2 & $+0.050$ & $+0.050$ & $+0.001$ & $+0.028$ & 0.842 \\
     &  & 3 & $+0.046$ & $+0.046$ & \boldmath$-0.001$ & $+0.028$ & 0.842 \\
    \midrule
    Ego-MC-Bench & Greedy & 0 & $+0.208$ & $+0.208$ & $+0.208$ & $+0.210$ & 1.000 \\
     &  & 1 & $+0.197$ & $+0.169$ & $+0.162$ & $+0.172$ & 0.946 \\
     &  & 2 & $+0.278$ & $+0.278$ & $+0.278$ & $+0.275$ & 1.000 \\
     &  & 3 & $+0.264$ & $+0.259$ & $+0.257$ & $+0.243$ & 0.994 \\
    \addlinespace[2pt]
     & Thresholded & 0 & $+0.190$ & $+0.178$ & $+0.175$ & $+0.171$ & 0.966 \\
     &  & 1 & $+0.082$ & $+0.008$ & \boldmath$-0.026$ & $+0.021$ & 0.552 \\
     &  & 2 & $+0.223$ & $+0.178$ & $+0.166$ & $+0.202$ & 0.992 \\
     &  & 3 & $+0.242$ & $+0.209$ & $+0.200$ & $+0.222$ & 0.996 \\
    \bottomrule
  \end{tabular*}
  \caption{\textbf{Margins of ECT against a stronger, learned video-blind policy}$^\dagger$. Bold margins fall inside the frontier.
  ``Rules'': the hand-written rules and their mixtures (\cref{tab:vertices}); the main text uses the second column. ``+ learned'': the upper hull of that frontier and of 18,900 learned blind policies, all tuned on the test set; ``paper's inputs'': instruction text, step index, elapsed time and training priors; ``all inputs'': in addition the position in the plan and the observed durations of the earlier steps of the recording.
  Held out: the blind policies on the hull of one half of the test recordings are scored on the other half, together with ECT; this frontier has no per-step mixtures.}
  \label{tab:blindstrong}
\end{table*}

\paragraph{Version of the Ego-MC-Bench baselines.}
The main paper takes the published per-step scores from Tables~2 and~3 of \texttt{arXiv:2606.09547v2}~\cite{bhattacharyya2026streaming}.
In v3, the same-backbone Qwen3-VL-2B (Ego-CoMist+) changes from 30.4 / 0.16 to 30.4 / 0.20 (IC-Acc / mistake F1), and the highest-F1 baseline, Qwen3.5-2B (Ego-CoMist+), changes from 37.1 / 0.20 to 33.0 / 0.27.
Comparing the 18 reported v3 scores with our frontiers, at least two lie above the rule frontier and none lies above the frontier of the main text, even when the learner uses only the paper's inputs. The published models were not rerun on our test recordings.

\paragraph{Why the clock scores.}
Over individual checks, raw elapsed time barely predicts the answer: its area under the ROC curve (AUROC) for the completion and the mistake question is 0.525 and 0.521 on the QICD test set, 0.558 and 0.500 on the validation split, and 0.342 and 0.432 on Ego-MC-Bench, where a longer elapsed time makes a positive answer less likely.
Elapsed time relative to the median training duration of the same instruction separates the completion answers better: AUROC 0.61 on the QICD test set and 0.69 on the validation split, with durations from the 209 training videos only (almost no Ego-MC-Bench instruction occurs in training).
The checks of the protocol thus carry a timing signal that a fine-tuned model can learn without the video. A negative drawn at the same elapsed time as its positive removes raw elapsed time from the pair, but only part of this signal, because the two checks usually belong to different instructions; \cref{sup:pairs} measures what remains.
Beyond this signal, the blind rules score through the sequential protocol.
The tolerance windows around annotated events cover 20.6\% of the stream time on the QICD test set and 28.9\% on Ego-MC-Bench.
In QICD, 91\% of the mistake feedback of the test set (89\% of the validation split) is annotated at the end of the step; in Ego-MC-Bench, 4\%.
On Ego-MC-Bench, 27 of the 264 mistake-free steps (10\%), 25 of them the last step of a session, carry only the instruction in the annotation, so their annotated end coincides with their start and any confirmation within 15\,s of the step start is credited; the zero-shot model is credited for 12 of them and ECT for 10--12.
Priors from the training split matter where recipes repeat: adding them to elapsed time raises the best blind F1 on QICD from 0.024 to 0.129 at IC-Acc $\geq 40$ and from 0.097 to 0.239 at IC-Acc $\geq 30$, and on Ego-MC-Bench from 0.107 to 0.130 at IC-Acc $\geq 40$.

\paragraph{GuideMe.}
For GuideMe~\cite{liu2026guideme} we re-read the 13 speaking-time results of its main table; we run no model on it.
We use the public test annotations (473 videos, 9,479 response points) and the official scoring script, and no video.
The sampling of silent points is not part of the released code and is reconstructed from the description in the paper; the reconstruction gives a silent share of 43.74\%, against the 43.6 reported for a system that never speaks.
The blind rules respond as a function of the time since the previous response, optionally with response-rate priors; since the training annotations are not public, the priors are cross-fitted on the test videos in five folds.
\Cref{tab:guideme} compares each published result with the blind policy at the same false-alarm rate.
Margins within about one point should be read as ``on the frontier''.
The scores of the benchmark's language-model judge are not reproduced.
A policy that never speaks obtains a score of 43.7, which matches or exceeds 11 of the 13 results.

\begin{table*}[t]
  \centering
  \small
  \setlength{\tabcolsep}{4pt}
  \resizebox{\ifdim\width>\linewidth\linewidth\else\width\fi}{!}{%
  \begin{tabular}{@{}lccc@{}}
    \toprule
    Published result & False-alarm rate & Response rate & Margin over time-only frontier \\
    \midrule
    Doubao-Seed-1.8 & 79.8 & 87.9 & -6.1 to -5.9 \\
    GPT-5.2 & 69.5 & 70.0 & -19.4 to -18.7 \\
    Gemini 3 Pro & 52.1 & 65.4 & -13.6 to -12.8 \\
    Gemini 3.1 Pro & 83.1 & 88.8 & -6.5 to -6.3 \\
    Qwen2.5-VL-7B & 56.3 & 62.8 & -18.8 to -18.1 \\
    Qwen3-VL-8B & 87.9 & 88.3 & -8.9 to -8.8 \\
    Qwen3-VL-30B-A3B & 84.5 & 89.5 & -6.4 to -6.2 \\
    Qwen3.5-397B-A17B & 18.3 & 33.4 & -21.5 to -20.1 \\
    VideoLLM-online & 100.0 & 100.0 & +0.0 to +0.0 \\
    Dispider & 0.0 & 0.2 & -33.1 to -33.0 \\
    LiveStar & 51.2 & 57.0 & -21.3 to -20.6 \\
    MMDuet2 & 0.5 & 0.3 & -34.9 to -34.5 \\
    Qwen3-VL-8B (FT on GuideMe-Train) & 55.7 & 63.1 & -18.2 to -17.4 \\
    \bottomrule
  \end{tabular}}
  \caption{GuideMe, speaking-time metrics: response rate of each published result minus that of a time-only blind policy at the same false-alarm rate (points; range over five random seeds of the reconstructed silent-point sampling). Negative: inside the frontier; zero: on it.}
  \label{tab:guideme}
\end{table*}

\section{Full results}
\label{sup:results}

\paragraph{Per seed.}
\Cref{tab:seeds} gives every seed of ECT with both decodings, and \cref{tab:ablseeds} every seed of the ablation.
Against the zero-shot model with the memory prompt, completion accuracy rises by 29.0 to 31.9 points on QICD (four seeds) and by 39.8 to 52.7 points on Ego-MC-Bench (four seeds), each with a 95\% interval far from zero (paired bootstrap over videos).
After Holm correction over all comparisons of the main table, the gain in mistake F1 on QICD is significant for seeds 1 and 2.
The gain is spread over the videos: the median completion accuracy per video is 0.33--0.44 on QICD for the four seeds of ECT and 0.00 for the zero-shot model, which confirms no step in time in 21 of the 35 videos that have mistake-free steps (ECT: 3--6); on Ego-MC-Bench the medians are 0.50--0.67 against 0.125 (17 of 39 videos without a timely confirmation, against 0--2).

\begin{table*}[t]
  \centering
  \small
  \setlength{\tabcolsep}{4pt}
  \resizebox{\ifdim\width>\linewidth\linewidth\else\width\fi}{!}{%
  \begin{tabular}{@{}lcccccccccc@{}}
    \toprule
    & \multicolumn{5}{c}{QICD} & \multicolumn{5}{c}{Ego-MC-Bench} \\
    \cmidrule(lr){2-6}\cmidrule(l){7-11}
    ECT, Qwen3-VL-2B & IC-Acc & P & R & F1 & $\Delta$ & IC-Acc & P & R & F1 & $\Delta$ \\
    \midrule
    Seed 0, greedy & 41.2 & 0.208 & 0.228 & 0.217 & $+0.062$ & 67.0 & 0.510 & 0.131 & 0.208 & $+0.208$ \\
    Seed 0, thresholds & 42.0 & 0.312 & 0.300 & 0.306 & $+0.165$ & 55.3 & 0.377 & 0.139 & 0.203 & $+0.190$ \\
    Seed 1, greedy & 39.9 & 0.261 & 0.312 & 0.284 & $+0.107$ & 54.2 & 0.349 & 0.169 & 0.228 & $+0.197$ \\
    Seed 1, thresholds & 41.2 & 0.353 & 0.297 & 0.322 & $+0.167$ & 48.1 & 0.315 & 0.159 & 0.211 & $+0.082$ \\
    Seed 2, greedy & 39.5 & 0.273 & 0.262 & 0.267 & $+0.083$ & 59.1 & 0.490 & 0.194 & 0.278 & $+0.278$ \\
    Seed 2, thresholds & 41.6 & 0.338 & 0.274 & 0.303 & $+0.155$ & 53.0 & 0.430 & 0.202 & 0.274 & $+0.223$ \\
    Seed 3, greedy & 42.4 & 0.258 & 0.194 & 0.221 & $+0.087$ & 55.7 & 0.384 & 0.209 & 0.271 & $+0.264$ \\
    Seed 3, thresholds & 44.1 & 0.269 & 0.221 & 0.242 & $+0.137$ & 53.8 & 0.405 & 0.214 & 0.280 & $+0.242$ \\
    \midrule
    Mean, greedy & 40.8 & 0.250 & 0.249 & 0.247 & $+0.085$ & 59.0 & 0.433 & 0.176 & 0.246 & $+0.237$ \\
    Mean, thresholds & 42.2 & 0.318 & 0.273 & 0.293 & $+0.156$ & 52.5 & 0.382 & 0.178 & 0.242 & $+0.184$ \\
    \bottomrule
  \end{tabular}}
  \caption{ECT per seed, with greedy decoding (main setting) and with the pre-registered decision thresholds. P / R: mistake precision / recall.}
  \label{tab:seeds}
\end{table*}

\begin{table*}[t]
  \centering
  \small
  \setlength{\tabcolsep}{4pt}
  \resizebox{\ifdim\width>\linewidth\linewidth\else\width\fi}{!}{%
  \begin{tabular}{@{}lcccccc@{}}
    \toprule
    Training & QICD & P / R & $\Delta$ & Ego-MC-Bench & P / R & $\Delta$ \\
    \midrule
    Plain fine-tuning & 52.1 / 0.000 & 0.000 / 0.000 & $+0.000$ & 60.2 / 0.000 & 0.000 / 0.000 & $+0.000$ \\
    No neg., re-balanced sampling & 27.7 / 0.321 & 0.257 / 0.426 & $+0.063$ & 47.7 / 0.318 & 0.335 / 0.302 & $+0.185$ \\
    No paired neg., seed 0 & 23.1 / 0.258 & 0.179 / 0.464 & $-0.025$ & 47.7 / 0.341 & 0.314 / 0.373 & $+0.208$ \\
    No paired neg., seed 1 & 29.8 / 0.254 & 0.179 / 0.437 & $+0.007$ & 48.9 / 0.333 & 0.337 / 0.330 & $+0.214$ \\
    + notdone, seed 0 & 31.1 / 0.275 & 0.211 / 0.395 & $+0.035$ & 56.8 / 0.253 & 0.325 / 0.207 & $+0.253$ \\
    + notdone, seed 1 & 32.8 / 0.271 & 0.198 / 0.433 & $+0.041$ & 51.1 / 0.314 & 0.371 / 0.272 & $+0.231$ \\
    + notdone, seed 2 & 31.1 / 0.366 & 0.299 / 0.471 & $+0.126$ & -- & -- & -- \\
    + notdone, seed 3 & 40.8 / 0.278 & 0.207 / 0.422 & $+0.116$ & -- & -- & -- \\
    + clean, seed 0 & 43.3 / 0.302 & 0.276 / 0.335 & $+0.183$ & 54.5 / 0.222 & 0.371 / 0.159 & $+0.195$ \\
    + clean, seed 1 & 43.3 / 0.270 & 0.266 / 0.274 & $+0.151$ & 54.2 / 0.222 & 0.464 / 0.146 & $+0.191$ \\
    ECT, seed 0 & 41.2 / 0.217 & 0.208 / 0.228 & $+0.062$ & 67.0 / 0.208 & 0.510 / 0.131 & $+0.208$ \\
    ECT, seed 1 & 39.9 / 0.284 & 0.261 / 0.312 & $+0.107$ & 54.2 / 0.228 & 0.349 / 0.169 & $+0.197$ \\
    ECT, seed 2 & 39.5 / 0.267 & 0.273 / 0.262 & $+0.083$ & 59.1 / 0.278 & 0.490 / 0.194 & $+0.278$ \\
    ECT, seed 3 & 42.4 / 0.221 & 0.258 / 0.194 & $+0.087$ & 55.7 / 0.271 & 0.384 / 0.209 & $+0.264$ \\
    \bottomrule
  \end{tabular}}
  \caption{The ablation of the paper per seed (greedy decoding). The two extra seeds of ``+ notdone'' were pre-registered for QICD only. Seed 0 of ``+ notdone'' was resumed from a checkpoint at step 2,300 of 2,400.}
  \label{tab:ablseeds}
\end{table*}

\paragraph{Sizes, doses and variants.}
\Cref{tab:other} collects the remaining single-seed runs.
The amount of clean negatives moves the operating point: fewer give a lower completion accuracy and a higher mistake F1, more the reverse.
The decomposition of \cref{tab:decomp-qicd} shows how: from 0.75 times to three times the clean negatives, the share of mistake-free steps voided by a false alarm falls from 22.3\% to 9.2\%, and the accuracy on mistake pairs rises from 0.732 to 0.770 (0.745--0.769 for the four seeds of ECT; \cref{tab:pairs}).
The amount of training data does not act monotonically: with 48 videos ECT is beyond the frontier on both benchmarks, and with 102 videos it is inside on QICD ($\Delta=-0.025$) and beyond on Ego-MC-Bench; with one seed each we draw no conclusion about data efficiency.
Two variants were evaluated on the test sets and not adopted: adding cross-recording pairs to the loss, and an auxiliary loss that aligns situations with feedback texts.
Both lie beyond the frontier; we did not adopt them because they add machinery without a consistent gain (the first has a significantly lower mistake F1 on Ego-MC-Bench).

\begin{table*}[t]
  \centering
  \small
  \setlength{\tabcolsep}{4pt}
  \resizebox{\ifdim\width>\linewidth\linewidth\else\width\fi}{!}{%
  \begin{tabular}{@{}lcccc@{}}
    \toprule
    Run (one seed each) & QICD & $\Delta$ & Ego-MC-Bench & $\Delta$ \\
    \midrule
    Qwen3-VL-4B, ECT, greedy & 40.3 / 0.227 & $+0.057$ & 61.7 / 0.175 & $+0.175$ \\
    Qwen3-VL-4B, ECT, thresholds & 44.1 / 0.306 & $+0.201$ & 56.4 / 0.205 & $+0.205$ \\
    InternVL3.5-2B, ECT & 47.9 / 0.256 & $+0.256$ & 64.0 / 0.104 & $+0.104$ \\
    InternVL3.5-2B, no paired neg. & 42.9 / 0.226 & $+0.100$ & 56.8 / 0.107 & $+0.107$ \\
    LLaVA-OneVision-7B, ECT & 49.6 / 0.171 & $+0.171$ & 64.0 / 0.090 & $+0.090$ \\
    \midrule
    ECT, clean negatives $\times 0.75$ & 37.8 / 0.272 & $+0.069$ & 59.5 / 0.229 & $+0.229$ \\
    ECT, clean negatives $\times 3$ & 52.5 / 0.222 & $+0.222$ & 60.2 / 0.209 & $+0.209$ \\
    ECT, 48 training videos & 45.8 / 0.138 & $+0.104$ & 54.2 / 0.169 & $+0.138$ \\
    ECT, 102 training videos & 28.2 / 0.230 & $-0.025$ & 56.4 / 0.170 & $+0.170$ \\
    \midrule
    Cross-recording pairs in the loss & 51.3 / 0.224 & $+0.224$ & 66.3 / 0.134 & $+0.134$ \\
    Auxiliary alignment loss, retrieval decoding & 45.0 / 0.263 & $+0.196$ & 62.9 / 0.184 & $+0.184$ \\
    \bottomrule
  \end{tabular}}
  \caption{Other backbones and sizes, the amount of clean negatives, the amount of training data, and two variants that we tried and did not adopt. IC-Acc / mistake F1 and margin over the frontier.}
  \label{tab:other}
\end{table*}

\paragraph{Confirmation time and step duration.}
On Ego-MC-Bench, the time at which ECT confirms a mistake-free step correlates with the annotated duration of the step: $r=0.687$, 0.449, 0.574 and 0.593 for the four seeds, against 0.102 for the zero-shot model. For the single-seed ablations cited in \cref{sec:evidence}, $r$ is 0.68 with notdone negatives alone and 0.30--0.40 for variants without them.
We do not use the same statistic on QICD, where the median training duration of the instruction alone correlates with the test duration at $r=0.444$.
Each run's correlation uses the steps it confirms without an alarm. On the 107 such steps shared by every run (four ECT seeds and two per ablation), the correlations are 0.735 for ECT and 0.831 for ``+ notdone'', compared with 0.606 for ``+ clean'' and 0.577 without paired negatives. The interval for the ECT versus ``+ clean'' difference spans zero ($+0.129$, 95\% interval $[-0.05, +0.37]$).

\begin{figure*}[t]
  \centering
  \includegraphics[width=\linewidth]{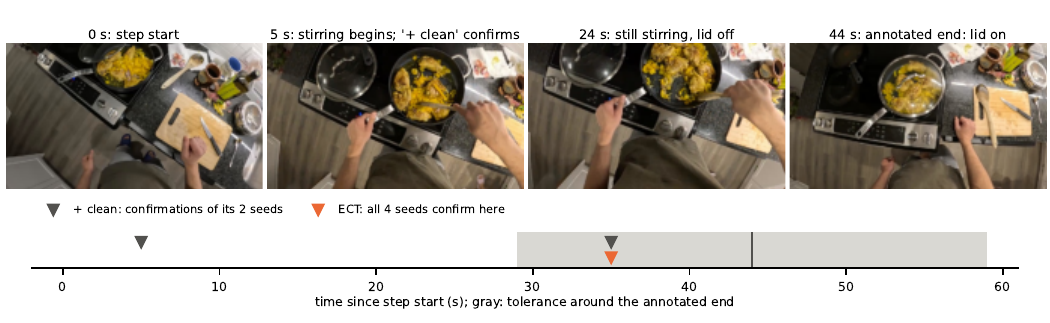}
  \caption{\textbf{An early confirmation that ECT avoids.} On this 44\,s Ego-MC-Bench step, seed 0 of ``+ clean'' confirms at the first check, 5\,s in, when stirring has just begun. All four seeds of ECT and seed 1 of ``+ clean'' confirm at 35\,s, within tolerance. The example has the median duration among audited steps without an explicit duration in the instruction where ECT confirms on time and ``+ clean'' confirms early.}
  \label{fig:notdone-case}
\end{figure*}

\paragraph{What the notdone negatives change$^\dagger$.}
On Ego-MC-Bench, the number of mistake-free steps confirmed more than 15\,s before their end is 37--42 for the four seeds of ECT and 55 for ``+ notdone'', against 62 for ``+ clean'' and 82 without paired negatives; on QICD the counts are 35--41 and 49, against 50 and 65 (75 with re-balanced sampling); the ablated variants have one seed each in these counts.
\Cref{tab:notdone} focuses on the longest third of the mistake-free steps, where early confirmations are most likely, with two seeds of each ablated variant.
On the 84 long steps of Ego-MC-Bench, ECT confirms on time 14.3 points more often than ``+ clean'' (95\% interval $[+5.8, +24.4]$), mostly because it confirms early less often ($-9.5$ points) and partly because it raises fewer alarms ($-6.0$, not significant); every seed of ECT (33--54\%) is above both seeds of ``+ clean'' (25 and 31\%).
The on-time gain persists on long steps whose instructions state no duration: $+12.5$ points $[+3.8, +21.6]$. Among steps whose instructions state a duration, early confirmation falls by 21.6 points $[-37.5, -6.7]$.
Steps of at least 20\,s confirmed at the very first check, 5\,s after their start, are 4.3--7.2\% for ECT against 9.4 and 12.2\% for ``+ clean'' ($-4.9$ points $[-7.1, -2.8]$).
With notdone negatives alone, 27.4\% of long steps are confirmed on time.
On QICD, whose test recipes all occur in training so that duration priors are informative, no such difference appears ($-1.9$ points $[-9.5, +4.5]$ on time).
\Cref{fig:notdone-case} illustrates the pattern with one step chosen by a fixed rule.

\begin{table*}[t]
  \centering\small\setlength{\tabcolsep}{5pt}
  \resizebox{\ifdim\width>\linewidth\linewidth\else\width\fi}{!}{%
  \begin{tabular}{llccccc}
    \toprule
    Outcome (\%) & Long steps ($n$) & ECT & + clean & + notdone & No neg. & ECT $-$ + clean [95\% CI] \\
    \midrule
    On time & Ego-MC-Bench (84) & 42.3 & 28.0 & 27.4 & 7.7 & $+14.3$ [+5.8, +24.4] \\
     & QICD (65) & 35.8 & 37.7 & 22.3 & 7.7 & $-1.9$ [-9.5, +4.5] \\
    \midrule
    Early & Ego-MC-Bench (84) & 40.5 & 50.0 & 48.2 & 57.1 & $-9.5$ [-18.4, -1.9] \\
     & QICD (65) & 33.8 & 36.2 & 27.7 & 26.9 & $-2.3$ [-10.7, +3.4] \\
    \midrule
    Alarm raised & Ego-MC-Bench (84) & 13.7 & 19.6 & 22.6 & 34.5 & $-6.0$ [-12.5, +0.3] \\
     & QICD (65) & 19.2 & 18.5 & 46.2 & 63.1 & $+0.8$ [-5.1, +9.0] \\
    \bottomrule
  \end{tabular}}
  \caption{\textbf{Long mistake-free steps}$^\dagger$ (the longest third of the steps scored in every run: at least 38.2\,s on Ego-MC-Bench and 51.8\,s on QICD), greedy decoding; mean over 4 seeds of ECT and 2 seeds of each variant; paired bootstrap over videos.}
  \label{tab:notdone}
\end{table*}

\paragraph{What the mistake F1 on Ego-MC-Bench rewards.}
Without paired negatives the mistake F1 on Ego-MC-Bench is higher than that of ECT.
\Cref{tab:f1null} asks where an F1 comes from, with two nulls that keep the alarms of a run.
Permuting the alarms across steps lowers the F1 of every fine-tuned model by a third or more, so these models pick steps that contain a mistake; the zero-shot 2B model does no better than its permuted alarms.
Moving the alarms to random checks inside their steps changes the F1 by at most 0.03: with a tolerance of $\pm$15\,s and free repeats inside the window, the metric hardly scores \emph{when} in a step an alarm is raised.
The variant without negatives reaches its higher F1 by raising more than twice as many alarms as ECT on average, in about twice as many steps and at lower precision (0.33 versus 0.43); the price is the completion accuracy, since every alarm in a mistake-free step voids it.

\begin{table*}[t]
  \centering
  \small
  \setlength{\tabcolsep}{5pt}
  \resizebox{\ifdim\width>\linewidth\linewidth\else\width\fi}{!}{%
  \begin{tabular}{@{}lccccccc@{}}
    \toprule
    Run (Ego-MC-Bench, greedy decoding) & Alarms & On mistake steps & P & R & F1 & F1, times shuffled [95\%] & F1, steps shuffled [95\%] \\
    \midrule
    No paired neg., seed 0 & 615 & 0.76 & 0.314 & 0.373 & 0.341 & 0.314 [0.294, 0.332] & 0.198 [0.169, 0.226] \\
    No paired neg., seed 1 & 517 & 0.77 & 0.337 & 0.330 & 0.333 & 0.312 [0.292, 0.330] & 0.188 [0.160, 0.217] \\
    No neg., re-balanced & 496 & 0.80 & 0.335 & 0.302 & 0.318 & 0.295 [0.275, 0.314] & 0.176 [0.147, 0.206] \\
    + notdone, seed 0 & 323 & 0.77 & 0.325 & 0.207 & 0.253 & 0.250 [0.227, 0.273] & 0.150 [0.117, 0.181] \\
    + clean, seed 0 & 214 & 0.67 & 0.371 & 0.159 & 0.222 & 0.202 [0.181, 0.221] & 0.135 [0.104, 0.169] \\
    ECT, seed 0 & 144 & 0.78 & 0.510 & 0.131 & 0.208 & 0.184 [0.162, 0.206] & 0.105 [0.079, 0.138] \\
    ECT, seed 1 & 243 & 0.64 & 0.349 & 0.169 & 0.228 & 0.211 [0.192, 0.231] & 0.146 [0.113, 0.177] \\
    ECT, seed 2 & 223 & 0.80 & 0.490 & 0.194 & 0.278 & 0.247 [0.224, 0.267] & 0.134 [0.104, 0.171] \\
    ECT, seed 3 & 316 & 0.76 & 0.384 & 0.209 & 0.271 & 0.263 [0.240, 0.285] & 0.141 [0.114, 0.169] \\
    Qwen3-VL-2B zero-shot & 459 & 0.56 & 0.200 & 0.232 & 0.215 & 0.271 [0.252, 0.292] & 0.209 [0.194, 0.227] \\
    Qwen3.8-Max zero-shot & 402 & 0.65 & 0.325 & 0.322 & 0.324 & 0.304 [0.276, 0.331] & 0.227 [0.194, 0.259] \\
    \bottomrule
  \end{tabular}}
  \caption{What the mistake F1 on Ego-MC-Bench rewards. ``On mistake steps'': share of the alarms raised in steps that contain a mistake (53\% of the steps do). ``Times shuffled'': the alarms stay in their steps and are moved to random checks of the step. ``Steps shuffled'': the alarm patterns are permuted across the steps of the same video. Means and 95\% ranges over 500 draws; our re-implementation of the official matching reproduces the counts of the official evaluation for every run.}
  \label{tab:f1null}
\end{table*}

\paragraph{Streaming protocol.}
\Cref{tab:stream} gives the streaming results that the paper summarizes.

\begin{table*}[t]
  \centering
  \small
  \setlength{\tabcolsep}{4pt}
  \resizebox{\ifdim\width>\linewidth\linewidth\else\width\fi}{!}{%
  \begin{tabular}{@{}lcc@{}}
    \toprule
    Model & QICD & Ego-MC-Bench \\
    \midrule
    ECT, seed 0 & 18.1 / 0.127 & 32.6 / 0.101 \\
    ECT, seed 1 & 16.4 / 0.142 & 24.2 / 0.115 \\
    ECT, seed 2 & 16.4 / 0.179 & 28.4 / 0.137 \\
    ECT, seed 3 & 22.3 / 0.175 & 27.3 / 0.170 \\
    + notdone & 13.4 / 0.129 & 23.9 / 0.128 \\
    No paired neg. & 8.0 / 0.105 & 11.7 / 0.125 \\
    Qwen3-VL-2B, zero-shot & 3.4 / 0.040 & 3.4 / 0.088 \\
    \midrule
    LiveMamba (published) & 12.6 / 0.19 & -- \\
    Streaming blind frontier, vertices & (0.0, 0.123), (1.3, 0.132), (3.8, 0.135), (13.9, 0.000) & (0.0, 0.115), (8.7, 0.000) \\
    \bottomrule
  \end{tabular}}
  \caption{Streaming protocol (no restart at ground-truth step starts): IC-Acc / mistake F1. The streaming blind frontier is low, because the errors of a blind rule accumulate across steps; the zero-shot model on Ego-MC-Bench and LiveMamba on QICD also lie beyond it, so we do not use it as evidence in the paper.}
  \label{tab:stream}
\end{table*}

\paragraph{QICD Main set.}
The paper uses the Advanced Planning set of QICD. \Cref{tab:mainset} repeats the evaluation on its Main set (109 test videos, none of them in our training or validation data; 36 of them are the recordings of the Advanced Planning test set, here with the plans of the Main set), under the per-step protocol with the official prompts and evaluation and both decodings, and compares with the results published on this set: those of the benchmark paper~\cite{bhattacharyya2025livemamba} and those reported for the benchmark's 2026 challenge~\cite{var2026aicoach}.
The video-blind frontier of this set is built as in \cref{sup:frontier}; its vertices are (16.5, 0.219), (18.7, 0.219), (27.4, 0.217), (32.9, 0.214), (37.6, 0.198), (47.3, 0.153), (51.0, 0.134), (54.1, 0.078), (56.3, 0.000)%
.
The general-purpose models and the second challenge entry lie inside it. Two published systems lie beyond it: LiveMamba ($+0.056$) and the challenge winner, a training-free pipeline of two proprietary models ($+0.306$).
ECT lies beyond it with nothing trained or tuned on the Main set: all four seeds with greedy decoding ($+0.072$ to $+0.181$), all four with the thresholds frozen on the Advanced Planning validation split ($+0.022$ to $+0.050$), and the recipe on Qwen3-VL-4B ($+0.151$), InternVL3.5-2B ($+0.103$) and LLaVA-OneVision-7B ($+0.017$).
The same fine-tuning without paired negatives (two seeds, $-0.032$ and $-0.046$) and all zero-shot models lie inside (from $-0.058$ for Qwen3.5-397B-A17B to $-0.217$ for InternVL3.5-2B).
With greedy decoding, the mistake F1 of ECT (0.192--0.224) is close to that of LiveMamba (0.19) and the second challenge entry (0.20) and below that of the winner (0.31), at a completion accuracy of 47.4--56.0 against 51.0, 31.4 and 56.2.
These runs were made after the challenge had closed and are not challenge entries.

\begin{table}[t]
  \centering
  \small
  \setlength{\tabcolsep}{4pt}
  \resizebox{\ifdim\width>\linewidth\linewidth\else\width\fi}{!}{%
  \begin{tabular}{@{}llcc@{}}
    \toprule
    System (QICD Main, 109 test videos) & Source & IC-Acc / F1 & $\Delta$ \\
    \midrule
    MR-CAS (winner) & 2026 challenge & 56.2 / 0.31 & $+0.306$ \\
    LiveMamba & benchmark paper & 51.0 / 0.19 & $+0.056$ \\
    Second entry & 2026 challenge & 31.4 / 0.20 & $-0.015$ \\
    VideoLLaMA3-7B & benchmark paper & 17.8 / 0.15 & $-0.069$ \\
    Qwen2-VL-7B & benchmark paper & 19.4 / 0.11 & $-0.109$ \\
    Qwen2.5-VL-7B & benchmark paper & 38.9 / 0.06 & $-0.132$ \\
    Qwen3-VL-8B baseline & 2026 challenge & 19.8 / 0.00 & $-0.219$ \\
    \midrule
    ECT, Qwen3-VL-2B, seed 0 & this work & 53.5 / 0.202 & $+0.113$ \\
    ECT, Qwen3-VL-2B, seed 1 & this work & 47.4 / 0.224 & $+0.072$ \\
    ECT, Qwen3-VL-2B, seed 2 & this work & 50.9 / 0.212 & $+0.077$ \\
    ECT, Qwen3-VL-2B, seed 3 & this work & 56.0 / 0.192 & $+0.181$ \\
    ECT, thresholded, seed 0 & this work & 41.2 / 0.203 & $+0.022$ \\
    ECT, thresholded, seed 1 & this work & 40.3 / 0.223 & $+0.038$ \\
    ECT, thresholded, seed 2 & this work & 41.7 / 0.229 & $+0.050$ \\
    ECT, thresholded, seed 3 & this work & 46.1 / 0.205 & $+0.046$ \\
    ECT, Qwen3-VL-4B & this work & 55.0 / 0.197 & $+0.151$ \\
    ECT, InternVL3.5-2B & this work & 52.6 / 0.208 & $+0.103$ \\
    ECT, LLaVA-OneVision-7B & this work & 52.7 / 0.120 & $+0.017$ \\
    No paired neg., seed 0 & this work & 27.2 / 0.185 & $-0.032$ \\
    No paired neg., seed 1 & this work & 27.2 / 0.171 & $-0.046$ \\
    Qwen3-VL-2B, zero-shot & this work & 4.5 / 0.047 & $-0.172$ \\
    Qwen3-VL-2B, zero-shot, memory prompt & this work & 5.0 / 0.097 & $-0.122$ \\
    InternVL3.5-2B, zero-shot & this work & 26.8 / 0.000 & $-0.217$ \\
    Qwen3.5-397B-A17B, zero-shot & this work & 29.7 / 0.158 & $-0.058$ \\
    \bottomrule
  \end{tabular}}
  \caption{The Main set of QICD under the per-step protocol (IC-Acc / mistake F1 and margin over the video-blind frontier of this set). Published values are as reported, with two decimals. Our models are those of the paper, trained on the Advanced Planning set, with nothing tuned on the Main set; the thresholded rows use the thresholds frozen on the Advanced Planning validation split.}
  \label{tab:mainset}
\end{table}

\IfFileExists{supp/tab_controls.tex}{\begin{table*}[t]
  \centering
  \small
  \setlength{\tabcolsep}{4pt}
  \resizebox{\ifdim\width>\linewidth\linewidth\else\width\fi}{!}{%
  \begin{tabular}{@{}lccccc@{}}
    \toprule
    Training (QICD test set) & IC-Acc / F1 & P / R & $\Delta$ & Pairs, completion & Pairs, mistake \\
    \midrule
    ECT, seed 0 & 41.2 / 0.217 & 0.208 / 0.228 & $+0.062$ & 0.903 & 0.766 \\
    ECT, seed 1 & 39.9 / 0.284 & 0.261 / 0.312 & $+0.107$ & 0.904 & 0.745 \\
    Random-moment negatives, seed 0 & 41.2 / 0.292 & 0.296 / 0.289 & $+0.137$ & 0.897 & 0.780 \\
    Random-moment negatives, seed 1 & 36.1 / 0.271 & 0.272 / 0.270 & $+0.059$ & 0.910 & 0.700 \\
    Trained and tested on black frames, seed 0 & 10.1 / 0.000 & 0.000 / 0.000 & $-0.316$ & 0.639 & 0.557 \\
    Trained and tested on black frames, seed 1 & 18.1 / 0.000 & 0.000 / 0.000 & $-0.307$ & 0.618 & 0.540 \\
    \bottomrule
  \end{tabular}}
  \caption{Two controls with the recipe of ECT and the same seeds: negatives drawn at random moments of the same steps instead of the matched elapsed time, and training and testing on black frames (a learned video-blind model).}
  \label{tab:controls}
\end{table*}
}{}
\paragraph{Is exact elapsed-time matching needed?}
\Cref{tab:controls} compares matched negatives with an equal number drawn at random moments within the same steps. On QICD, relative to ECT with the same seed, IC changes by $+0.0$ $[-5.3, +4.9]$ and $-3.8$ $[-10.1, +2.7]$ points, and mistake F1 by $+0.075$ $[+0.021, +0.128]$ and $-0.013$ $[-0.071, +0.035]$ (paired 95\% video-bootstrap intervals, seeds 0 and 1). Accuracy on mistake pairs is 0.780 and 0.700, versus 0.766 and 0.745 for matched ECT. Elapsed time alone gives AUROC 0.555--0.561 for completion and 0.607--0.620 for mistakes on the random-moment examples, versus 0.541 and 0.606 with matching: within a recording, random moments leave elapsed time almost as uninformative as matched ones.

\emph{Transfer to Ego-MC-Bench.} We evaluated the same two existing random-moment adapters on the 39 released videos, without training or tuning for this test. Their official IC/F1 is 59.5/0.272 and 57.6/0.261. Both IC values exceed the maximum IC of the canonical blind family, so the corresponding margins are $+0.272$ and $+0.261$. We specified the same-seed comparisons before these runs and resampled the same 39 videos for each pair (2,000 draws, bootstrap seed 0). In seed 0, random rather than matched moments reduce IC by $7.6$ points (signed difference $[-12.2,-3.2]$) but raise F1 by $0.063$ $[+0.013,+0.109]$. In seed 1, the differences are $+3.4$ points $[-2.2,+8.4]$ and $+0.033$ $[-0.003,+0.074]$. For on-time confirmation on the 282 mistake-then-success steps, the corresponding differences are $-2.5$ points $[-7.1,+1.5]$ and $+1.8$ $[-3.0,+6.9]$ (all 282 reached in all four runs; not official IC).

Both kinds of same-recording negatives put the model beyond the frontier; the choice moves the IC/F1 trade-off on Ego-MC-Bench in seed 0, in a direction that seed 1 does not repeat. The variant in \cref{tab:other} adds cross-recording pairs to the loss.

\paragraph{Completion after a mistake$^\dagger$.}
Official IC excludes steps containing mistakes. To check whether the timing behavior also appears on such steps, we re-read frozen predictions on the 282 Ego-MC-Bench steps with an annotated success after the last annotated mistake. The first confirmation is on time if it falls strictly within 15\,s of the last success; for the fine-tuned runs below all 282 steps were reached. Averaged over seeds, the rates are 56.4\% for ECT (four seeds), 53.7\% for ``+ clean'', 54.1\% for ``+ notdone'', and 42.0\% with no paired negatives (two seeds each). Relative to ``+ clean'', ECT has fewer early confirmations ($-5.1$ points $[-7.7,-2.6]$) but more late or absent confirmations ($+2.5$ $[+1.1,+3.9]$); the net on-time gain is $+2.7$ $[+0.4,+5.0]$ (paired 95\% video-bootstrap intervals). The planned short-step check did not pass: late or absent confirmations also increased for short steps. This retrospective endpoint does not explain the difference between the two benchmarks.

The text-duration heuristic reaches 39.4\%, with its delays chosen on this same cohort; only 14 instructions explicitly state a time. On InternVL3.5-2B (one seed), ECT and no paired negatives reach 49.3\% and 35.8\%, respectively, on the same 282 steps.
\paragraph{A video-blind model trained with our recipe.}
The last two rows of \cref{tab:controls} train and test the recipe of ECT on black frames, so that the model has only the instruction, the length of the window and what it learned in training.
Both seeds stay inside the frontier: they never raise an alarm and reach a completion accuracy of 10.1 and 18.1 (mean 14.1, as reported in the main text), where the hand-written rules reach a mistake F1 of 0.316 and 0.307.
On the time-matched pairs they score 0.62--0.64 and 0.54--0.56, the part above 0.50 coming from the instruction text, against 0.90 and 0.75--0.77 for ECT with the same seeds.
Under the same training recipe, the black-frame models do not reproduce the performance of ECT with real video.

\section{Decoding, thresholds and operating points}
\label{sup:operating}

\paragraph{Thresholded decoding.}
Before any test run of ECT we pre-registered a variant that reads the probability of ``yes'' at the first answer token and compares it with two thresholds chosen on the validation split by a fixed rule (0.2 for completion, 0.35 for mistakes), and froze it.
Greedy decoding is the main setting because it is deterministic and needs no validation data; it became the main setting after both decodings, run in the same frozen batch, had been evaluated on the test sets, and both are reported for every seed.
\Cref{tab:seeds} reports both for every seed: thresholds help every seed on QICD and lower completion accuracy on Ego-MC-Bench.

\paragraph{Shared validation alarm cap$^\dagger$.}
We select each variant's threshold on the validation split under the same cap as thresholded ECT: 432 raw alarms. Using scored alarms for both ECT and the variants (a cap of 387) selects the same thresholds. \Cref{tab:budget} replays the test runs from their logged probabilities at these thresholds, without matching test-set alarm counts.
At this cap, the two variants without clean negatives have the largest QICD IC shortfalls relative to thresholded ECT seed 0 ($-7.1$ and $-5.9$ points; both 95\% intervals just exclude zero). Their intervals for QICD F1 and for both Ego-MC-Bench metrics span zero.
With clean negatives alone (``+ clean''), IC is 2.9 points lower on QICD and 3.0 points lower on Ego-MC-Bench; both intervals span zero. On Ego-MC-Bench its mistake F1 is higher by $+0.058$ $[+0.012, +0.104]$.
These are single-seed comparisons with unadjusted video-bootstrap intervals; \cref{tab:budget} reports equivalence tests separately. The long-step analysis in \cref{sup:results} asks how adding notdone negatives to a recipe with clean negatives changes early confirmations on Ego-MC-Bench.

The shared validation cap is a secondary operating-point comparison, separate from the official main-table scores.

\begin{table*}[t]
  \centering
  \small
  \setlength{\tabcolsep}{4pt}
  \resizebox{\ifdim\width>\linewidth\linewidth\else\width\fi}{!}{%
  \begin{tabular}{@{}llccccccc@{}}
    \toprule
    Benchmark & Variant & $\tau$ & Val.\ alarms & IC-Acc / F1 & $\Delta$IC-Acc [95\% CI] & Equiv. & $\Delta$F1 [95\% CI] & Equiv. \\
    \midrule
    QICD & No paired neg. & 0.60 & 409 & 34.9 / 0.285 & -7.1 [-14.4, -0.4] & no & -0.021 [-0.082, +0.033] & no \\
    QICD & + notdone & 0.47 & 412 & 36.1 / 0.311 & -5.9 [-11.0, -0.5] & no & +0.005 [-0.043, +0.058] & yes \\
    QICD & + clean & 0.33 & 409 & 39.1 / 0.342 & -2.9 [-7.9, +1.8] & no & +0.036 [-0.016, +0.088] & no \\
    \midrule
    Ego-MC-Bench & No paired neg. & 0.60 & 409 & 52.3 / 0.248 & -3.0 [-8.4, +3.1] & no & +0.045 [-0.011, +0.101] & no \\
    Ego-MC-Bench & + notdone & 0.47 & 412 & 54.5 / 0.188 & -0.8 [-5.3, +3.7] & yes & -0.014 [-0.059, +0.025] & no \\
    Ego-MC-Bench & + clean & 0.33 & 409 & 52.3 / 0.261 & -3.0 [-7.3, +1.2] & no & +0.058 [+0.012, +0.104] & no \\
    \bottomrule
  \end{tabular}}
  \caption{Shared validation alarm cap. $\tau$ is the lowest threshold on a 0.01 grid with at most 432 raw validation alarms. Thresholded ECT (seed 0): 42.0 / 0.306 on QICD and 55.3 / 0.203 on Ego-MC-Bench. Differences are variant minus ECT (one seed each); 95\% intervals use a paired video bootstrap without adjustment for multiple comparisons. Equiv.: equivalence test with pre-declared bounds of $\pm5$ IC-Acc and $\pm0.05$ F1.}
  \label{tab:budget}
\end{table*}

\begin{figure*}[t]
  \centering
  \includegraphics[width=\linewidth]{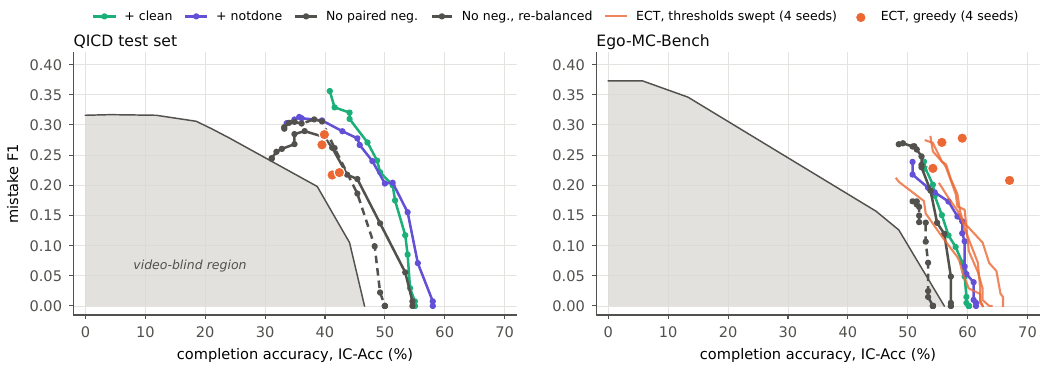}
  \caption{\textbf{Threshold sweeps.} Each curve sweeps the alarm threshold of one thresholded model on the test set (one seed per ablated variant); dots are ECT with greedy decoding. The sweeps of ECT itself can be replayed on Ego-MC-Bench only. On QICD the ablated variants pass close to ECT's operating point and separate at high completion accuracy; on Ego-MC-Bench the curves of ECT lie to the right of the others.}
  \label{fig:sweep}
\end{figure*}

\paragraph{Sweeping the threshold.}
\Cref{tab:sweep} and \cref{fig:sweep} sweep the alarm threshold of each thresholded model on the test set.
On QICD, a thresholded model without paired negatives reaches similar IC but lower F1 than thresholded ECT seed 0: 41.2 / 0.262 at a threshold of 0.75, versus 42.0 / 0.306 for ECT.
The variants separate where completion accuracy is high, and in the limit without alarms, where the variants with notdone negatives confirm more steps in time.
These sweeps map the trade-off (thresholds swept on the test set, one seed per variant).

\begin{table*}[t]
  \centering
  \small
  \setlength{\tabcolsep}{4pt}
  \resizebox{\ifdim\width>\linewidth\linewidth\else\width\fi}{!}{%
  \begin{tabular}{@{}lcccccc@{}}
    \toprule
    & \multicolumn{3}{c}{QICD} & \multicolumn{3}{c}{Ego-MC-Bench} \\
    \cmidrule(lr){2-4}\cmidrule(l){5-7}
    Variant (one seed each) & IC, no alarms & F1 at IC $\geq 40$ & F1 at IC $\geq 50$ & IC, no alarms & F1 at IC $\geq 40$ & F1 at IC $\geq 50$ \\
    \midrule
    ECT, seed 0 & -- & -- & -- & 65.9 & 0.203 & 0.203 \\
    ECT, seed 1 & -- & -- & -- & 62.5 & 0.211 & 0.174 \\
    ECT, seed 2 & -- & -- & -- & 62.5 & 0.274 & 0.274 \\
    ECT, seed 3 & -- & -- & -- & 64.0 & 0.280 & 0.280 \\
    ECT, Qwen3-VL-4B & 62.2 & 0.306 & 0.286 & 67.0 & 0.205 & 0.205 \\
    + clean & 55.0 & 0.356 & 0.200 & 60.2 & 0.239 & 0.239 \\
    + notdone & 58.0 & 0.289 & 0.204 & 61.4 & 0.239 & 0.239 \\
    No paired neg. & 54.6 & 0.262 & 0.056 & 57.2 & 0.270 & 0.265 \\
    No neg., re-balanced sampling & 50.0 & 0.262 & 0.000 & 54.2 & 0.173 & 0.173 \\
    \bottomrule
  \end{tabular}}
  \caption{Sweeping the alarm threshold of each thresholded model on the test set (exact offline replay of the logged probabilities). ``IC, no alarms'': completion accuracy when no alarm is raised; the other columns give the best mistake F1 among operating points with at least that completion accuracy. The QICD logs of the 2B ECT runs cannot be replayed exactly from these logs, so those cells are empty (\cref{sup:decomp} replays seed 0 on QICD from per-check logs recorded later).}
  \label{tab:sweep}
\end{table*}

\paragraph{Use of the test sets.}
We ran each configuration once per test set, except for the duplicate zero-shot queries noted in \cref{sup:protocol}.
Before the configuration of ECT was frozen, the zero-shot models and the plain fine-tuning recipe of the ablation had already been evaluated on both test sets.
ECT was then selected and frozen on the validation split alone; its test runs, the three further seeds, the ablations, the other backbones and sizes, and the variants of \cref{tab:other} were registered before they were run. Greedy decoding was chosen as the main setting after both decodings had been run on the test sets.
The runs added later were registered in the same way before they were run: the two further seeds of ``+ notdone'', the random-moment negatives, the models trained on black frames, the step-aligned swap and the evaluation on the Main set (both decodings), on which nothing was tuned.
The Ego-MC-Bench evaluation of the existing random-moment adapters used a separately specified comparison with same-seed ECT (\cref{sup:results}). The reach and mistake-then-success diagnostics, stronger learned blind policy, shared alarm-cap comparison, long-step timing analysis and case review were designed or revised after the original test results were available, and so were the clock analyses of \cref{sup:frontier,sup:decomp,sup:perturb} (relative-duration AUROC, protocol replay, zero-shot alarms, time-stamp probe) and the use of the learned blind policy in the frontier of the main text. For the replay we logged the probabilities of both questions at every check of the QICD test set for two frozen models (ECT seed 0 and the variant without paired negatives); these runs replay or perturb frozen models. They do not select new model parameters or change the primary endpoint.
Both pre-registered decodings were run in the same frozen batch and are reported for every seed.

\section{Time-matched pairs}
\label{sup:pairs}

\begin{figure}[t]
  \centering
  \includegraphics[width=\linewidth]{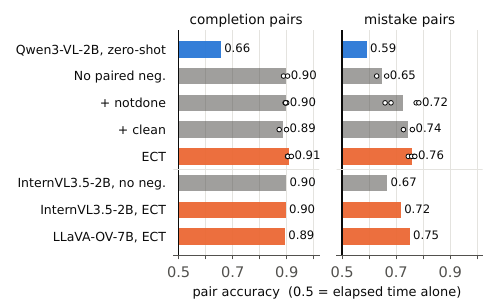}
  \caption{\textbf{ECT ranks checks taken at the same elapsed time.}
  Accuracy on 931 time-matched test pairs; bars start at 0.5, the score of any policy of elapsed time alone; ranking by the instruction's typical duration scores 0.65 on completion and 0.54 on mistake pairs. Dots are individual seeds.
  Fine-tuning on the protocol's checks is enough for completion pairs; paired negatives add the gain on mistake pairs, on two backbones (\cref{tab:pairs} has all models).}
  \label{fig:pairs}
\end{figure}

\paragraph{Construction and results.}
A completion pair is a check at which a step has ended and a check of another step of the same recording at the same elapsed time that still has at least 20\,s to run; a mistake pair is a check at an annotated mistake and a check of a mistake-free step of the same recording at the same elapsed time with at least 15\,s to run.
The QICD test videos give 417 completion pairs and 514 mistake pairs; in three pairs the two windows differ in length by more than one frame because the video ends.
\Cref{tab:pairs} gives all models and \cref{fig:pairs} the summary.
The two checks of a pair share the elapsed time and the window length but usually not the instruction (they share it in 35 of the completion and 56 of the mistake pairs). A blind policy that ranks them by elapsed time over the median training duration of their instruction scores 0.651 on completion pairs and 0.539 on mistake pairs, and the recipe of ECT trained on black frames 0.62--0.64 and 0.54--0.56 (\cref{sup:results}); ECT scores 0.90--0.92 and 0.75--0.77.
Differences between groups of models, with 95\% intervals from a paired bootstrap over videos:
ECT against the three models without paired negatives (two seeds and the re-balanced variant), $+0.020$ $[-0.002, +0.046]$ on completion pairs and $+0.104$ $[+0.044, +0.156]$ on mistake pairs;
``+ clean'' and ``+ notdone'' against the same three models on mistake pairs, $+0.088$ $[+0.027, +0.146]$ and $+0.061$ $[+0.024, +0.099]$ (two seeds each);
ECT against ``+ clean'' on completion pairs, $+0.022$ $[+0.004, +0.040]$.
On the side of the pair without evidence, ECT gives ``yes'' a probability above 0.5 in 6--9\% of the completion pairs and 4--8\% of the mistake pairs, against 20--22\% and 23--32\% for the three models without them.
In QICD most mistake feedback is annotated at the end of the step, so a mistake pair usually also differs in how close the step is to its end: for 490 of the 505 distinct positives fewer than 15\,s of the step remain, while the negative side has at least 15\,s to run by construction.
As a control we compare, on the fixed validation situations of \cref{tab:probe}, the probability of ``yes'' to the mistake question at annotated mistakes with that at the completion of mistake-free steps, within 10\,s bins of elapsed time, so that both sides lie at a step end.
The AUROC is 0.57--0.64 for ECT (seeds 0--2), 0.65 for ``+ clean'', 0.73 for ``+ notdone'' and 0.74 without paired negatives (ECT minus the last, 95\% interval $[-0.18, -0.09]$).
Together with \cref{tab:probe}, this suggests that ECT's gain on mistake pairs reflects fewer alarms while steps are still running, rather than better discrimination at the step end.
The price is visible in the ``both right'' column: ECT ranks mistake pairs better but answers ``yes'' less often on the side with the mistake, which is the low mistake recall discussed in the paper.

\begin{table*}[t]
  \centering
  \small
  \setlength{\tabcolsep}{4pt}
  \resizebox{\ifdim\width>\linewidth\linewidth\else\width\fi}{!}{%
  \begin{tabular}{@{}lcccc@{}}
    \toprule
    & \multicolumn{2}{c}{Completion pairs (417)} & \multicolumn{2}{c}{Mistake pairs (514)} \\
    \cmidrule(lr){2-3}\cmidrule(l){4-5}
    Model & Pair accuracy [95\% CI] & Both right & Pair accuracy [95\% CI] & Both right \\
    \midrule
    Qwen3-VL-2B, zero-shot & 0.657 [0.595, 0.720] & 0.412 & 0.592 [0.494, 0.683] & 0.119 \\
    \midrule
    ECT, seed 0 & 0.903 [0.856, 0.950] & 0.607 & 0.766 [0.692, 0.833] & 0.173 \\
    ECT, seed 1 & 0.904 [0.844, 0.953] & 0.640 & 0.745 [0.667, 0.816] & 0.249 \\
    ECT, seed 2 & 0.904 [0.861, 0.946] & 0.631 & 0.757 [0.683, 0.824] & 0.175 \\
    ECT, seed 3 & 0.917 [0.874, 0.953] & 0.564 & 0.769 [0.696, 0.838] & 0.161 \\
    \midrule
    No paired neg., seed 0 & 0.888 [0.845, 0.928] & 0.602 & 0.628 [0.568, 0.693] & 0.304 \\
    No paired neg., seed 1 & 0.903 [0.859, 0.942] & 0.635 & 0.665 [0.600, 0.736] & 0.267 \\
    No neg., re-balanced sampling & 0.871 [0.826, 0.915] & 0.578 & 0.671 [0.612, 0.731] & 0.294 \\
    + notdone, seed 0 & 0.896 [0.845, 0.946] & 0.633 & 0.659 [0.593, 0.732] & 0.282 \\
    + notdone, seed 1 & 0.900 [0.850, 0.948] & 0.633 & 0.774 [0.720, 0.825] & 0.327 \\
    + notdone, seed 2 & 0.894 [0.843, 0.942] & 0.667 & 0.784 [0.726, 0.838] & 0.323 \\
    + notdone, seed 3 & 0.896 [0.846, 0.942] & 0.626 & 0.681 [0.629, 0.738] & 0.274 \\
    + clean, seed 0 & 0.899 [0.863, 0.935] & 0.638 & 0.727 [0.648, 0.805] & 0.265 \\
    + clean, seed 1 & 0.872 [0.821, 0.919] & 0.600 & 0.760 [0.680, 0.830] & 0.206 \\
    \midrule
    ECT, clean negatives $\times 0.75$ & 0.902 [0.849, 0.949] & 0.588 & 0.732 [0.649, 0.810] & 0.193 \\
    ECT, clean negatives $\times 3$ & 0.909 [0.870, 0.944] & 0.628 & 0.770 [0.691, 0.844] & 0.144 \\
    ECT, 48 training videos & 0.881 [0.818, 0.943] & 0.561 & 0.706 [0.625, 0.781] & 0.128 \\
    ECT, 102 training videos & 0.866 [0.804, 0.920] & 0.511 & 0.732 [0.662, 0.797] & 0.249 \\
    \midrule
    InternVL3.5-2B, ECT & 0.897 [0.852, 0.941] & 0.633 & 0.718 [0.642, 0.792] & 0.195 \\
    InternVL3.5-2B, no paired neg. & 0.899 [0.854, 0.944] & 0.667 & 0.668 [0.602, 0.737] & 0.220 \\
    LLaVA-OneVision-7B, ECT & 0.894 [0.848, 0.936] & 0.583 & 0.751 [0.673, 0.821] & 0.086 \\
    \bottomrule
  \end{tabular}}
  \caption{Time-matched test pairs, all models. Pair accuracy: share of pairs in which the correct side gets the higher probability of ``yes'' (any policy of elapsed time alone: 0.5; elapsed time relative to the instruction's median training duration: 0.651 and 0.539). Both right: share of pairs in which the model would answer ``yes'' on the positive side and ``no'' on the other at a threshold of 0.5.}
  \label{tab:pairs}
\end{table*}

\paragraph{The example of \cref{fig:teaser}.}
The pair was chosen by a rule fixed in advance: completion pairs whose two windows differ by at most one frame, that all four ECT seeds rank correctly and that the zero-shot model ranks wrongly or ties, ordered by the mean margin of ECT; we show the median pair.
\Cref{tab:fig1cands} lists it with its three neighbours on either side.

\begin{table*}[t]
  \centering
  \small
  \setlength{\tabcolsep}{4pt}
  \resizebox{\ifdim\width>\linewidth\linewidth\else\width\fi}{!}{%
  \begin{tabular}{@{}lcccc@{}}
    \toprule
    Recording & Elapsed time (s) & Zero-shot & No paired neg. & ECT (4 seeds) \\
    \midrule
    2\_47 & 64.4 & 0.76 / 0.98 & 0.78 / 0.71 & 0.56 / 0.11 \\
    25\_48 & 36.1 & 0.14 / 0.73 & 0.98 / 0.88 & 0.94 / 0.47 \\
    15\_46 & 91.4 & 0.72 / 0.99 & 0.96 / 0.15 & 0.50 / 0.01 \\
    10\_48 (\cref{fig:teaser}) & 20.5 & 0.28 / 0.82 & 0.84 / 0.32 & 0.69 / 0.20 \\
    26\_46 & 41.8 & 0.11 / 0.17 & 0.97 / 0.62 & 0.68 / 0.20 \\
    8\_50 & 27.9 & 0.01 / 0.23 & 0.05 / 0.01 & 0.50 / 0.01 \\
    17\_43 & 34.5 & 0.08 / 0.33 & 0.71 / 0.01 & 0.50 / 0.00 \\
    \bottomrule
  \end{tabular}}
  \caption{The completion pair of \cref{fig:teaser} and the three pairs on either side of it in the ranking of the selection rule. Probability of ``yes'' on the finished step / on the unfinished step.}
  \label{tab:fig1cands}
\end{table*}

\begin{figure*}[t]
  \centering
  \includegraphics[width=\linewidth]{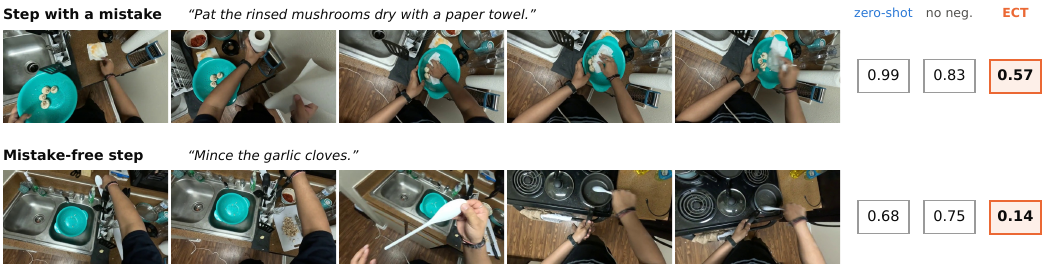}
  \caption{\textbf{The one mistake pair that passes the strict rule.} Both checks are made 20.5\,s into their step. The mistake annotation indicates that the mushrooms were crushed rather than patted dry. Boxes: probability of ``yes'' to the mistake question (no neg.: mean of three models without paired negatives; ECT: mean of four seeds).}
  \label{fig:mistake-pair}
\end{figure*}

\paragraph{Examples for the mistake question are rare.}
We looked for a mistake pair to show in the same way: all four seeds rank the pair correctly and all three models without paired negatives rank it wrongly or tie.
\Cref{tab:miscands} lists the pairs nearest the median.
In most of them ECT answers ``no'' on both sides and merely ranks them correctly.
Under the stricter requirement that ECT answers ``yes'' on the step with the mistake and ``no'' on the other while a model without negatives raises a false alarm, one of the 931 pairs qualifies (the first row of \cref{tab:miscands}; \cref{fig:mistake-pair}).

\begin{table*}[t]
  \centering
  \small
  \setlength{\tabcolsep}{4pt}
  \resizebox{\ifdim\width>\linewidth\linewidth\else\width\fi}{!}{%
  \begin{tabular}{@{}llccc@{}}
    \toprule
    Recording & Step with the mistake & Zero-shot & No paired neg. & ECT (4 seeds) \\
    \midrule
    20\_48 & pat the rinsed mushrooms dry with a paper towel & 0.99 / 0.68 & 0.83 / 0.75 & 0.57 / 0.14 \\
    3\_49 & measure 1/8 teaspoon of baking powder and add it to the mu & 0.96 / 0.73 & 0.52 / 0.73 & 0.47 / 0.19 \\
    2\_47 & microwave the plate, covered, on high for 1.5 minutes & 0.90 / 0.94 & 0.17 / 0.41 & 0.37 / 0.09 \\
    7\_135 & extract and add contents of an egg to a microwave-safe bow & 0.92 / 0.94 & 0.67 / 0.82 & 0.55 / 0.19 \\
    26\_46 & measure and add a pinch of salt to the mixing bowl & 0.95 / 0.68 & 0.23 / 0.51 & 0.45 / 0.01 \\
    17\_45 & add 1/4 teaspoon of salt to the bowl & 0.99 / 0.92 & 0.14 / 0.61 & 0.20 / 0.02 \\
    17\_45 & chop or grate the cucumber & 0.95 / 0.44 & 0.10 / 0.33 & 0.11 / 0.02 \\
    \bottomrule
  \end{tabular}}
  \caption{Mistake pairs that all four ECT seeds rank correctly and all three models without paired negatives rank wrongly or tie. Probability of ``yes'' on the step with the mistake / on the mistake-free step.}
  \label{tab:miscands}
\end{table*}

\section{Decision decomposition}
\label{sup:decomp}

\Cref{tab:decomp-qicd,tab:decomp-mc} give the decomposition behind the corresponding figure of the paper for more models.
On QICD, up to 4.6\% of the mistake-free steps fall in a category that the figure merges into ``not reached'': the instruction issued at that step does not match the ground truth.
The last two columns show the two sides of the operating point of ECT: fewer false alarms per mistake-free step, and fewer steps with a mistake flagged.
Plain fine-tuning, which never raises an alarm, confirms more mistake-free steps in time than ECT on both benchmarks (52.1 and 60.2, against 40.8 and 59.0; on QICD so do ``+ clean'' with 43.3 and tripled clean negatives with 52.5), so completion accuracy is read together with the mistake F1 it leaves.

\begin{table*}[t]
  \centering
  \small
  \setlength{\tabcolsep}{4pt}
  \resizebox{\ifdim\width>\linewidth\linewidth\else\width\fi}{!}{%
  \begin{tabular}{@{}lcccccccccc@{}}
    \toprule
    Model & $n$ & In time & Voided & Early & Late & Never & Not reached & Mismatch & Alarms & Flagged \\
    \midrule
    Qwen3-VL-2B, zero-shot & 1 & 10.5 & 56.3 & 4.6 & 3.4 & 0.0 & 22.3 & 2.9 & 0.73 & 68.6 \\
    Qwen3.8-Max, zero-shot & 1 & 23.9 & 52.9 & 16.4 & 0.4 & 0.4 & 1.3 & 4.6 & 0.99 & 64.0 \\
    Plain fine-tuning & 1 & 52.1 & 0.0 & 34.5 & 8.4 & 0.0 & 0.4 & 4.6 & 0.00 & 0.0 \\
    No neg., re-balanced sampling & 1 & 27.7 & 39.5 & 21.8 & 2.5 & 1.3 & 2.9 & 4.2 & 0.92 & 55.5 \\
    No paired neg. & 2 & 26.5 & 47.9 & 17.6 & 1.2 & 0.2 & 1.9 & 4.6 & 1.39 & 64.4 \\
    + notdone & 4 & 34.0 & 39.9 & 15.0 & 5.6 & 0.3 & 1.0 & 4.3 & 1.12 & 57.0 \\
    + clean & 2 & 43.3 & 25.4 & 16.2 & 9.2 & 0.9 & 0.7 & 4.4 & 0.61 & 40.9 \\
    ECT & 4 & 40.8 & 26.2 & 13.2 & 12.5 & 0.9 & 2.0 & 4.4 & 0.56 & 34.5 \\
    Random-moment negatives & 2 & 38.7 & 31.1 & 13.0 & 10.9 & 0.2 & 1.6 & 4.4 & 0.62 & 36.6 \\
    ECT, clean negatives $\times 0.75$ & 1 & 37.8 & 22.3 & 14.7 & 14.3 & 0.8 & 5.9 & 4.2 & 0.43 & 33.5 \\
    ECT, clean negatives $\times 3$ & 1 & 52.5 & 9.2 & 18.1 & 15.1 & 0.0 & 0.4 & 4.6 & 0.17 & 19.1 \\
    InternVL3.5-2B, no paired neg. & 1 & 42.9 & 26.9 & 21.8 & 3.8 & 0.0 & 0.0 & 4.6 & 0.85 & 42.4 \\
    InternVL3.5-2B, ECT & 1 & 47.9 & 20.2 & 16.8 & 6.7 & 0.0 & 4.2 & 4.2 & 0.54 & 35.6 \\
    ECT, video swapped (seed 0) & 1 & 13.4 & 19.3 & 13.4 & 28.2 & 5.0 & 16.4 & 4.2 & 0.47 & 26.7 \\
    \bottomrule
  \end{tabular}}
  \caption{Decision decomposition on the QICD test set: outcome of every mistake-free step under official scoring (\%; means over $n$ seeds). Mismatch: the instruction given at that step does not match the ground truth.  Alarms: false alarms per mistake-free step. Flagged: share of steps with a mistake on which the model raises an alarm (\%).}
  \label{tab:decomp-qicd}
\end{table*}

\begin{table*}[t]
  \centering
  \small
  \setlength{\tabcolsep}{4pt}
  \resizebox{\ifdim\width>\linewidth\linewidth\else\width\fi}{!}{%
  \begin{tabular}{@{}lcccccccccc@{}}
    \toprule
    Model & $n$ & In time & Voided & Early & Late & Never & Not reached & Mismatch & Alarms & Flagged \\
    \midrule
    Qwen3-VL-2B, zero-shot & 1 & 14.4 & 73.1 & 7.6 & 0.0 & 0.0 & 4.9 & 0.0 & 0.77 & 84.2 \\
    Qwen3.8-Max, zero-shot & 1 & 37.9 & 39.4 & 15.5 & 6.1 & 0.4 & 0.8 & 0.0 & 0.53 & 61.4 \\
    No paired neg. & 2 & 48.3 & 25.0 & 25.0 & 1.5 & 0.0 & 0.2 & 0.0 & 0.50 & 57.4 \\
    + notdone & 2 & 54.0 & 17.2 & 21.4 & 6.7 & 0.4 & 0.4 & 0.0 & 0.34 & 38.3 \\
    + clean & 2 & 54.4 & 15.2 & 21.0 & 8.5 & 0.2 & 0.8 & 0.0 & 0.24 & 23.1 \\
    ECT & 4 & 59.0 & 14.0 & 15.9 & 9.3 & 0.8 & 1.1 & 0.0 & 0.23 & 26.6 \\
    InternVL3.5-2B, ECT & 1 & 64.0 & 4.2 & 22.7 & 8.3 & 0.0 & 0.8 & 0.0 & 0.11 & 12.4 \\
    \bottomrule
  \end{tabular}}
  \caption{Decision decomposition on the Ego-MC-Bench: outcome of every mistake-free step under official scoring (\%; means over $n$ seeds). Mismatch: the instruction given at that step does not match the ground truth.  Alarms: false alarms per mistake-free step. Flagged: share of steps with a mistake on which the model raises an alarm (\%).}
  \label{tab:decomp-mc}
\end{table*}

\paragraph{Which question carries the gain?$^\dagger$}
For ECT and the variant without paired negatives (seed 0 of each), we logged the probability of ``yes'' to both questions at every check of the QICD test set and replayed the official per-step loop offline under the thresholded decoding of \cref{sup:operating}.
The replay matches the official IC-Acc of the thresholded runs (42.0 and 31.1) and approximates their mistake F1 (0.309 and 0.256 against 0.306 and 0.244); 11 and 12 steps are still unconfirmed 60\,s after their annotated end, where the log stops, and are counted as confirmed at the next check (ending the recording there instead gives 34.0 and 25.2).
Replacing the mistake probabilities of the variant by those of ECT, with its own completion probabilities, raises its IC-Acc from 31.1 to 40.8, close to the 42.0 of ECT: most of the gain in completion accuracy comes from fewer false alarms. At thresholds of 0.5 on both questions, the closest to greedy decoding, the same exchange raises it from 23.1 to 38.7, against 39.1 for ECT (more steps outlast the log there: 39--54).
Over the checks of mistake-free steps, 7.3\% raise an alarm for ECT and 18.3\% for the variant, and 26.5\% and 40.8\% of these steps are voided.
With every alarm removed, ECT still confirms more steps in time than the variant at every completion threshold from 0.1 to 0.6 (at best 61.3 against 54.6), so the two also differ in when they confirm, beyond the threshold; at higher thresholds a growing share of steps outlasts the logged checks.

\paragraph{Why the zero-shot model raises alarms$^\dagger$.}
On the fixed validation situations of \cref{tab:probe}, zero-shot Qwen3-VL-2B (one check with the official prompt) answers ``yes'' to the mistake question with a mean probability of 0.795 where there is no mistake and 0.798 where there is one.
With black frames the first falls to 0.146, and with up to 60\,s of elapsed time added (the window padded with copies of its first frame and cut to 240 frames) it stays at 0.729 (0.678 on the windows that receive the full 60\,s): the model responds to a person at work, not to the mistake, and its alarms do not grow with elapsed time.
Under the per-step protocol this voids steps from the start: in 83\% of the QICD test steps in which the zero-shot model with the memory prompt raises a false alarm, the first one comes at the first check, 5\,s in (90\% on Ego-MC-Bench), against 0--9\% for the four ECT seeds.

\section{What the feedback says}
\label{sup:content}

The benchmarks credit a detection by its time alone.
To see what the feedback says, we compared 25 randomly drawn detections of thresholded ECT (seed 0) on the QICD test set with the annotated feedback by hand: 9 name the annotated mistake, 3 name it in part, and 13 name a different mistake.
The 13 often use a generic phrase, such as a spill or a wrong amount; the alarm comes at the right time, but the model does not identify what went wrong.
Recall also differs across categories (\cref{tab:category}): mistakes of timing and temperature, which leave little visual trace within a window, are rarely detected.
The detections are timely: the median offset between an alarm and the annotated mistake is $-0.4$\,s on QICD and $-1.0$\,s on Ego-MC-Bench.

\begin{table}[t]
  \centering
  \small
  \setlength{\tabcolsep}{4pt}
  \resizebox{\ifdim\width>\linewidth\linewidth\else\width\fi}{!}{%
  \begin{tabular}{@{}lcc@{}}
    \toprule
    Mistake category & Detected / annotated & Recall \\
    \midrule
    Preparation & 28 / 68 & 0.41 \\
    Measurement & 27 / 74 & 0.36 \\
    Technique & 25 / 85 & 0.29 \\
    Timing & 3 / 26 & 0.12 \\
    Temperature & 1 / 10 & 0.10 \\
    \bottomrule
  \end{tabular}}
  \caption{Mistake recall by category on the QICD test set (ECT, seed 0, greedy decoding; a mistake counts as detected if any alarm falls within 15\,s of it).}
  \label{tab:category}
\end{table}

\begin{figure*}[t]
  \centering
  \includegraphics[width=\linewidth]{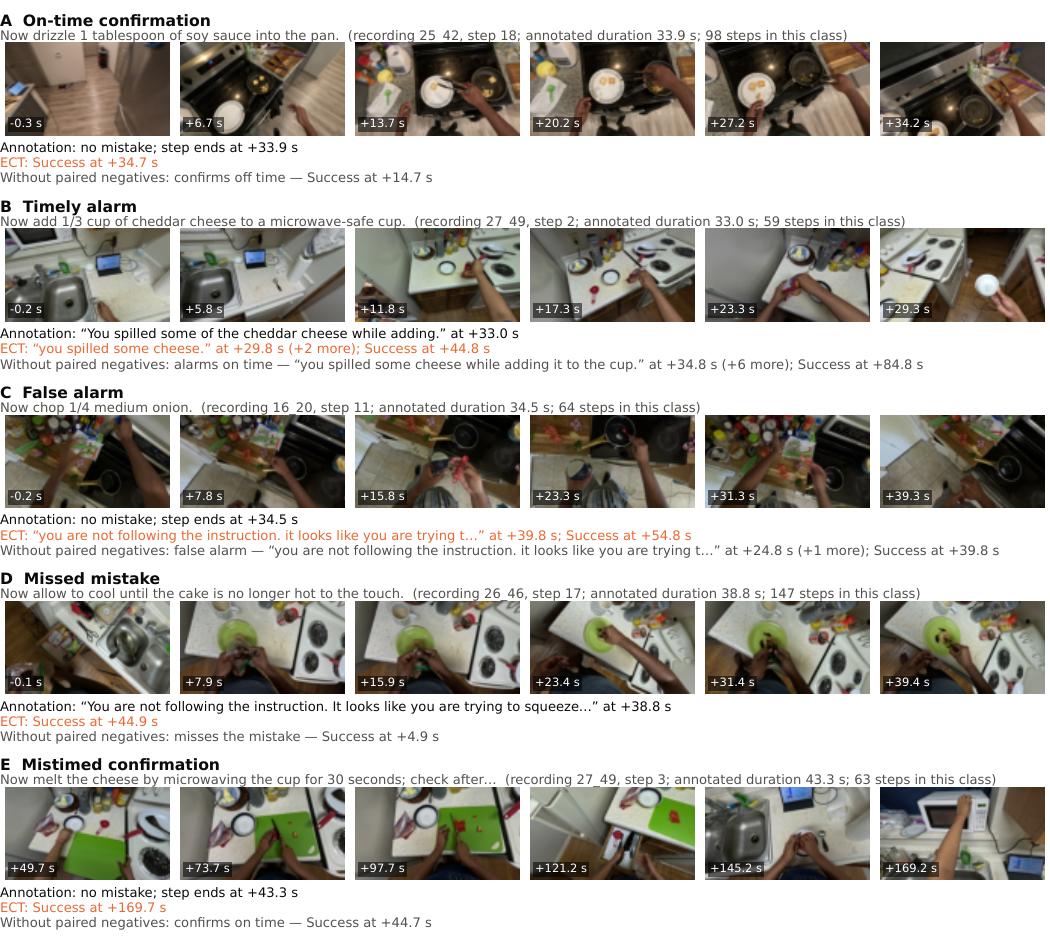}
  \caption{\textbf{Success and failure cases on QICD.} Each row shows six uniformly spaced frames of ECT's input window at the decision shown, the annotation and outputs from ECT and the model trained without paired negatives on the same step. Times are relative to the step start.}
  \label{fig:qual}
\end{figure*}

\paragraph{Success and failure cases.}
\Cref{fig:qual} shows five outcomes of ECT (seed 0, greedy decoding) on the QICD test set, with one step per outcome chosen by a fixed rule before any frame was viewed: the step with the median annotated duration within that outcome.
Over the test set, ECT confirms 98 mistake-free steps on time and confirms 63 others too early (34) or too late (29); it raises a false alarm in 64 mistake-free steps; it alarms on time in 59 steps with a mistake and stays silent in 147. The remaining 30 steps with a mistake receive alarms only outside the tolerance. Thirteen mistake-free steps have no scored decision (11 end-of-video segments with empty annotated instructions and two steps after the last prediction); they remain in the official IC denominator.
In the failure cases, ECT alarms 5\,s after the annotated end of a mistake-free step (C), confirms 6\,s after an annotated mistake without saying anything (D), and confirms 126\,s after the annotated end of a step (E).
On case A, the model trained without paired negatives confirms 19\,s early. On case B, it alarms seven times, compared with three alarms from ECT.

\section{Video perturbations}
\label{sup:perturb}

\paragraph{All perturbations.}
\Cref{tab:perturb-full} gives the validation perturbations for all variants. On the QICD test set, swapping the recording for ECT seed 0 reduces IC-Acc / mistake F1 from 41.2 / 0.217 to 13.4 / 0.094, as reported in \cref{sec:evidence}; \cref{tab:decomp-qicd} breaks down the test outcomes.
The last two perturbations of ECT change how the window is presented (32 frames sampled from it, or only its last 30\,s) while the model was trained on the full window since the step start; they lower completion accuracy.
A swapped video is another recording of the same recipe; for the 6 of the 36 test videos and 3 of the 57 validation videos whose recipe has no second recording in the split, it is a recording of another recipe.

\begin{table}[t]
  \centering
  \small
  \setlength{\tabcolsep}{4pt}
  \resizebox{\ifdim\width>\linewidth\linewidth\else\width\fi}{!}{%
  \begin{tabular}{@{}lccc@{}}
    \toprule
    Input (QICD validation split) & IC-Acc & F1 & Alarms raised \\
    \midrule
    ECT, seed 0 & 47.2 & 0.154 & 328 \\
    \quad frames shuffled & 37.1 & 0.091 & 353 \\
    \quad another recording & 14.7 & 0.067 & 376 \\
    \quad last frame repeated & 0.7 & 0.009 & 2 \\
    \quad black frames & 0.0 & 0.000 & 0 \\
    \quad 32 frames, uniformly sampled & 15.0 & 0.061 & 669 \\
    \quad window fixed to the last 30\,s & 21.4 & 0.086 & 224 \\
    \midrule
    ECT, seed 1 & 41.5 & 0.168 & 414 \\
    \quad another recording & 13.7 & 0.070 & 444 \\
    + clean & 40.7 & 0.203 & 415 \\
    \quad another recording & 12.8 & 0.078 & 473 \\
    + notdone & 38.1 & 0.202 & 636 \\
    \quad another recording & 10.4 & 0.075 & 792 \\
    No paired neg. & 23.6 & 0.163 & 1045 \\
    \quad another recording & 8.9 & 0.070 & 1376 \\
    \quad black frames & 1.5 & 0.054 & 118 \\
    \midrule
    Qwen3-VL-2B, zero-shot (memory prompt) & 3.8 & 0.075 & 685 \\
    \quad another recording & 0.3 & 0.047 & 600 \\
    \bottomrule
  \end{tabular}}
  \caption{Video perturbations on the QICD validation split, with the same protocol and instructions in every row. Swapping the video sharply lowers completion accuracy even without paired negatives. On black frames ECT raises no alarm, while the variant without paired negatives raises 118 (109 false). The zero-shot model's low completion accuracy on the original video leaves little room for a further drop.}
  \label{tab:perturb-full}
\end{table}

\begin{figure*}[t]
  \centering
  \includegraphics[width=\linewidth]{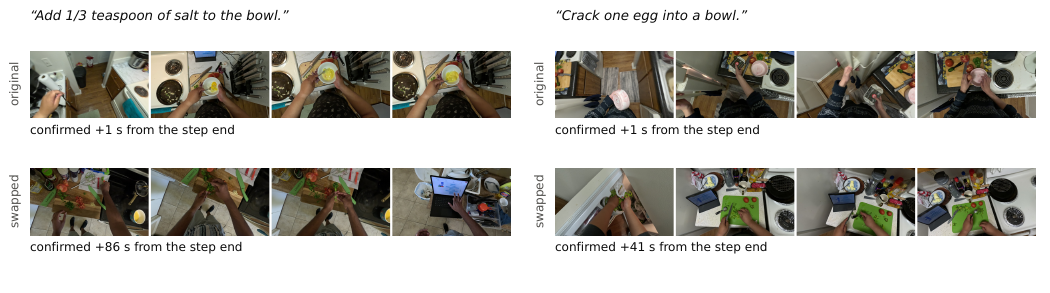}
  \caption{\textbf{Two further swap candidates} (rows 1 and 3 of \cref{tab:swapcands}): frames of the original and swapped recordings at the same indices, with the time at which ECT (seed 0) confirms the step.}
  \label{fig:swap-more}
\end{figure*}

\paragraph{The swap example.}
The step shown in \cref{fig:swap} was chosen as follows.
Among the mistake-free test steps that ECT (seed 0, the only seed with a swapped test run) confirms in time under the original video, 28 receive no alarm and no confirmation within tolerance after the swap; we order them by duration and take the seven nearest the median (\cref{tab:swapcands}).
Among those seven, the main figure shows the example where the two recordings were easiest to tell apart visually; \cref{fig:swap-more} shows two others.

\begin{table*}[t]
  \centering
  \small
  \setlength{\tabcolsep}{4pt}
  \resizebox{\ifdim\width>\linewidth\linewidth\else\width\fi}{!}{%
  \begin{tabular}{@{}llccc@{}}
    \toprule
    Recording $\to$ swapped in & Step & Duration (s) & Confirmed, original (s) & Confirmed, swapped (s) \\
    \midrule
    22\_40 $\to$ 22\_41 & crack one egg into a bowl & 29 & +1 & +41 \\
    25\_48 $\to$ 25\_42 & pat the tofu dry with a towel & 32 & +3 & +28 \\
    16\_44 $\to$ 16\_20 & add 1/3 teaspoon of salt to the bowl & 33 & +1 & +86 \\
    7\_50 $\to$ 7\_135 & add 1/2 tablespoon of sweet and sour sauce to the bo & 35 & +4 & +64 \\
    22\_41 $\to$ 22\_40 & stop stirring when it's nearly cooked to allow the e & 38 & +2 & +17 \\
    1\_43 $\to$ 1\_49 & place the egg from the cup over the lettuce & 41 & -11 & never \\
    16\_20 $\to$ 16\_44 (\cref{fig:swap}) & whisk the egg mixture in the bowl & 41 & -1 & +129 \\
    \bottomrule
  \end{tabular}}
  \caption{The seven candidates for the swap example of the paper: confirmation time relative to the annotated end of the step, under the original video and under the swapped one.}
  \label{tab:swapcands}
\end{table*}

\paragraph{Swapping in the same step of another recording.}
The swap above replaces frames index by index, so it changes both the visual evidence and its alignment with the instruction.
In a further run (ECT seed 0, QICD test set) every check instead receives the frames of the same instruction step in another recording of the same recipe, aligned at the step start.
Of the 193 steps that have such a counterpart, 37 (in 17 videos) are free of mistakes in both recordings, and 35 of these are confirmed.
Their confirmation time correlates with the duration of the step in the recording that the model sees at $r=0.38$ (95\% interval $[0.11, 0.69]$) and with its duration in the original recording at $r=0.21$ ($[-0.11, 0.47]$); with so few steps the difference is not significant ($[-0.24, 0.72]$).

\paragraph{Separability on fixed situations.}
\Cref{tab:probe} measures how well the probability of ``yes'' separates positive situations from silent ones on fixed validation situations.
For the mistake question, the variants without clean negatives separate these situations better than ECT; for the variant without paired negatives, the next paragraph shows that this advantage does not survive a change of the time stamps.

\begin{table}[t]
  \centering
  \small
  \setlength{\tabcolsep}{4pt}
  \resizebox{\ifdim\width>\linewidth\linewidth\else\width\fi}{!}{%
  \begin{tabular}{@{}lccc@{}}
    \toprule
    Model & $n$ & AUROC, completion & AUROC, mistake \\
    \midrule
    Qwen3-VL-2B, zero-shot & 1 & 0.750 & 0.566 \\
    ECT & 4 & 0.956 & 0.843 \\
    + clean & 1 & 0.963 & 0.863 \\
    + notdone & 1 & 0.968 & 0.937 \\
    No paired neg. & 2 & 0.975 & 0.951 \\
    No neg., re-balanced sampling & 1 & 0.983 & 0.974 \\
    \bottomrule
  \end{tabular}}
  \caption{Separability of the first-token probability of ``yes'' on 1,233 fixed validation situations (positive situations against silence). With the original time stamps the variants without clean negatives separate mistakes \emph{better} than ECT on this probe; the time-stamp probe below shows that, for the variant without paired negatives, this advantage does not survive a change of the stamps. The negatives move the operating point.}
  \label{tab:probe}
\end{table}

\paragraph{Does the model read the clock?$^\dagger$}
On 411 of these situations we changed only the time stamps that the processor writes between the frames, doubling or halving every stamp; the pixels, the number of frames and the prompt are unchanged.
ECT (seed 0) barely responds: its AUROC is 0.945, 0.942 and 0.943 for completion and 0.841, 0.820 and 0.846 for mistakes with the original, doubled and halved stamps, and its probability of ``yes'' moves by 0.013--0.025 on average.
The mistake AUROC of the variant without paired negatives (seed 0) falls from 0.971 to 0.79--0.80 and that of plain fine-tuning from 0.961 to 0.60--0.61; the recipe of ECT trained and tested on black frames (\cref{sup:results}), which separates these situations at 0.920 for completion and 0.958 for mistakes without seeing any video, falls to 0.74--0.77 and 0.59--0.69.
Doubling and halving the stamps both raise the probability of ``yes'' on silent situations, by 0.14--0.16 for the variant without paired negatives, while that of ECT changes by at most 0.01: the variant reacts to stamps unlike those of training rather than reading their value. With rescaled stamps its mistake AUROC is no longer above that of ECT (ECT minus the variant $-0.130$ $[-0.190, -0.079]$ with the original stamps, $+0.025$ $[-0.031, +0.080]$ doubled and $+0.042$ $[-0.011, +0.100]$ halved; 95\% intervals over videos).
ECT does use how long a step usually takes: on the QICD test set its confirmation time follows the median training duration of the instruction beyond the duration of the step itself (partial correlation of log times 0.55--0.63 over four seeds, 0.48--0.54 without paired negatives, 0.63--0.64 for the two black-frame models). Its training pairs leave this cue in place: re-simulated from the annotations, the instruction's typical duration (from the other training videos) separates the two sides of 80--81\% of the completion pairs and 58--59\% of the mistake pairs, although elapsed time alone ties them.
On the time-matched pairs, elapsed time alone scores 0.5, the instruction's typical duration 0.65 and 0.54, and the recipe trained on black frames 0.62--0.64 and 0.54--0.56; ECT uses such information too, but scores 0.90--0.92 and 0.75--0.77 (\cref{sup:pairs}), and in this probe it does not respond to the time stamps.

\section{External systems}
\label{sup:external}

\Cref{tab:external} places two published streaming assistants~\cite{zhang2025proactive,chen2024videollmonline} (for the latter, the 8B v1+ checkpoint) in the per-step protocol on the QICD validation split, forcing a response at every check rather than letting the assistants choose when to speak.
Both lie inside the frontier and react little to black frames.
ProAssist raises 8,191 scored mistake alarms over the 57 videos, 38 of them credited, and 9,697 on black frames; VideoLLM-online never answers ``yes'' to either question, with or without the video.
Forced to answer, the two systems thus sit near the two blind extremes of the paper: alarms regardless of the video, and silence.

\begin{table}[t]
  \centering
  \small
  \setlength{\tabcolsep}{4pt}
  \resizebox{\ifdim\width>\linewidth\linewidth\else\width\fi}{!}{%
  \begin{tabular}{@{}lcccc@{}}
    \toprule
    System (QICD validation split) & Video & Alarms & Black frames & Alarms \\
    \midrule
    ProAssist & 4.4 / 0.009 & 8,191 & 0.2 / 0.004 & 9,697 \\
    VideoLLM-online & 0.0 / 0.000 & 0 & 0.0 / 0.000 & 0 \\
    \bottomrule
  \end{tabular}}
  \caption{Two published streaming assistants placed in the per-step protocol and forced to answer every check (IC-Acc / mistake F1, and the number of mistake alarms that the official evaluation scores as true or false positives over the 57 videos). This differs from their own setting, in which they decide when to speak.}
  \label{tab:external}
\end{table}

\section{Training data and recipe}
\label{sup:training}

\paragraph{Situations.}
From the 209 training videos we build 18,406 situations (seed 0): 1,938 mistake, 4,563 success, 6,315 silence, 363 shifted, 4,068 notdone and 1,159 clean.
Mistake situations are placed at the annotated feedback time and 5 and 10\,s later, success situations at the step end and 5\,s later, and silence situations every 20\,s inside a step, away from any annotated event.
A shifted situation is a check before an annotated mistake and outside its tolerance window, answered ``NO''.
Each training step draws 8 situations by kind with fixed weights: mistake 0.25, success 0.20, silence, notdone and clean 0.15 each, and shifted 0.10.

\paragraph{Optimization.}
We train LoRA adapters of rank 64 on Qwen3-VL-2B with AdamW, a learning rate of $2\times10^{-4}$ and cross-entropy on the answer tokens for 2,400 steps (19,200 situations, about one epoch). A run takes about 15 hours on one A100: five later runs of the same pipeline took 13.6--16.8 hours; the four seeds of ECT shared their GPUs with other jobs and took 18.0--28.6 hours of wall-clock time, seed 3 on an A100-80GB (peak memory 9.1\,GiB).
Seed 0 of ``+ notdone'' was interrupted and resumed from its checkpoint at step 2,300.

\paragraph{The three controls.}
The rows of the ablation differ from ECT as follows.
\emph{No paired negatives} differs in the negatives alone: the situations are built in the same way, the notdone and clean situations are removed, and the weights of the remaining kinds are renormalized (mistake 0.36, success 0.29, silence 0.21, shifted 0.14).
\emph{No negatives, re-balanced sampling} uses the same situations as the previous variant with other weights (silence 0.40, success 0.25, mistake 0.25, shifted 0.10), which brings the share of ``YES'' answers to the mistake question close to that of ECT.
\emph{Plain fine-tuning} is an earlier recipe and differs in four ways: one mistake situation per annotated mistake and one success situation per step (9,632 situations: 6,315 silence, 2,303 success, 652 mistake, 362 shifted); uniform sampling over situations, so that mistakes make up about 7\% of the draws; the other question asked on 30\% of the positive situations instead of 20\%; and no paired negatives.
It is trained for two epochs (2,408 steps) with the same model and optimizer.
With so few positive examples for the mistake question it learns never to raise an alarm, which under the official scoring voids no step and gives it the highest completion accuracy; the gap between it and ECT cannot be attributed to the negatives.

\paragraph{Splits and shortcuts.}
The training, validation and test videos of QICD Advanced Planning (209, 57 and 36) are disjoint, and Ego-MC-Bench was recorded separately.
All 18 recipes of the QICD test set occur in the training split, by design of the benchmark; 2 of the 562 instructions of Ego-MC-Bench do.
A classifier that reads only the instruction text separates the labels of our training situations with an AUROC of 0.553.
The same classifier reaches 0.918 when the synthetic counterfactual pairs of \citet{bhattacharyya2026streaming} are mixed in, because their rewritten instructions differ in wording from the original ones; this is one reason why we train on QICD alone.
A model fine-tuned with these pairs shows the shortcut directly: when the leading ``Now'' is removed from the instruction and the frames are left unchanged, its mean probability of ``yes'' to the mistake question rises from 0.017 by 0.73 over 240 validation situations (by at least 0.67 for each kind of situation), whereas the zero-shot model and a model trained on QICD alone change by 0.001.
Elapsed time separates the labels with an AUROC of 0.57--0.60 for the completion question and 0.55--0.56 for the mistake question over all training situations, and with 0.541 and 0.606 over the positives and their paired negatives.

\section{Code and data}
\label{sup:code}

\Cref{tab:vertices} lists the rule-based frontier vertices. Besides the aggregate results, this preprint documents the evaluation protocol and training recipe and illustrates selected decisions. This source package does not include the datasets, raw prediction logs, model weights or evaluation code. Access to the underlying datasets is governed by their respective licenses.

\paragraph{Acknowledgments.}
We used Anthropic Claude (Opus and Fable via Claude Code) for research code, experiment execution and analysis, literature review, and drafting text, tables and figures. OpenAI Codex assisted with preparing and checking the preprint. The authors take responsibility for the work.

\end{document}